\RequirePackage{fix-cm}
\documentclass[twocolumn]{svjour3}          
\smartqed  
\usepackage{xcolor}
\usepackage{tabularx}
\usepackage{multirow}
\usepackage{graphicx}
\usepackage{makecell}
\usepackage{amsmath}
\usepackage{color}
\usepackage{float}
\usepackage{rotating}
\usepackage{placeins}
\usepackage{amsfonts}
\usepackage{bm}

\def\eg{\textit{e.g.}}
\def\ie{\textit{i.e.}}

\usepackage[title]{appendix}

\usepackage[round,authoryear,comma,sort&compress,numbers]{natbib} 
\usepackage[colorlinks=true,linkcolor=blue,citecolor=black,urlcolor=blue]{hyperref}

\usepackage{diagbox}
\usepackage{cancel} 
\usepackage{mathrsfs} 
\usepackage{eqnarray} 

\usepackage{tikz}
\usepackage{ctable} 

\usepackage{algorithm}
\usepackage{algpseudocode}%
\usepackage{listings}%
\usepackage{bbding}
\usepackage{bbm}
\usepackage[caption=false, font=footnotesize]{subfig}
\usepackage{fontawesome}
\usepackage{array}

\usepackage{colortbl}  
\usepackage{arydshln}
\usepackage{longtable}
\usepackage{wrapfig}
\usepackage{bbold}
\usepackage{caption}
\usepackage{pifont}

\usepackage{xcolor}
\usepackage{hyperref}

\hypersetup{hidelinks,
	colorlinks=true,
	allcolors=black,
	pdfstartview=Fit,
	breaklinks=true}

\definecolor{lime}{HTML}{A6CE39}
\DeclareRobustCommand{\orcidicon}{
\begin{tikzpicture}
\draw[lime, fill=lime] (0,0)
circle[radius=0.13]
node[white]{{\fontfamily{qag}\selectfont \tiny \.{I}D}};
\end{tikzpicture}
\hspace{-2mm}
}
\foreach \x in {A, ..., Z}{%
\expandafter\xdef\csname orcid\x\endcsname{\noexpand\href{https://orcid.org/\csname orcidauthor\x\endcsname}{\noexpand\orcidicon}}
}

\newcommand{\PreserveBackslash}[1]{\let\temp=\\#1\let\\=\temp}
\newcolumntype{C}[1]{>{\PreserveBackslash\centering}p{#1}}
\newcolumntype{R}[1]{>{\PreserveBackslash\raggedleft}p{#1}}
\newcolumntype{L}[1]{>{\PreserveBackslash\raggedright}p{#1}}

\def\eg{\textit{e.g.}}
\def\ie{\textit{i.e.}}

\definecolor{battleshipgrey}{rgb}{0.52, 0.52, 0.51}
\definecolor{capri}{rgb}{0.0, 0.75, 1.0}
\definecolor{mediumspringgreen}{rgb}{0.0, 0.98, 0.6}
\definecolor{Gray}{rgb}{0.7,0.7,0.7}
\newcommand{\model}{Efficient4D}

\usepackage{makecell}
\usepackage{arydshln} 
\usepackage{pifont}
\usepackage{ulem}
\usepackage{tabularx}

\RequirePackage{xspace}
\makeatletter
\DeclareRobustCommand\onedot{\futurelet\@let@token\@onedot}
\def\@onedot{\ifx\@let@token.\else.\null\fi\xspace}
\def\eg{\textit{e.g}\onedot} 

\def\ie{\textit{i.e}\onedot}

\journalname{%
  \parbox[t]{8cm}{%
    International Journal of Computer Vision
  }%
}
\begin{document}

\title{\model{}: Fast Dynamic 3D Object Generation from a Single-view Video
}


\author{
	Zijie Pan$^1$ \and
	Zeyu Yang$^1$ \and
    Xiatian Zhu$2$ \and
    Li Zhang$^1$\hspace{-2mm}\orcidA{}
}




\institute{
	Corresponding author: Li Zhang  \at
             \email{lizhangfd@fudan.edu.cn}          \\
$^1$School of Data Science, Fudan University \\
$^2$University of Surrey
}
\date{13th November 2025}

\maketitle

\begin{abstract}
Generating dynamic 3D object from a single-view video is challenging due to the lack of 4D labeled data. 
An intuitive approach is to 
extend previous image-to-3D pipelines by transferring off-the-shelf image generation models such as score distillation sampling.
However, this approach would be slow and expensive to scale due to the need for back-propagating the information-limited supervision signals through a large pretrained model.
To address this, we propose an efficient video-to-4D object generation framework called {\bf\model{}}.
It generates high-quality spacetime consistent images under different camera views, and then uses them as labeled data to directly reconstruct the 4D content through a 4D Gaussian splatting model.
Importantly, our method can achieve real-time rendering under continuous camera trajectories. To enable robust reconstruction under sparse views, we introduce inconsistency-aware confidence-weighted loss design, along with a lightly weighted score distillation loss.
Extensive experiments on both synthetic and real videos show that \model{} offers a remarkable 10-fold increase in speed when compared to prior art alternatives while preserving the quality of novel view synthesis. For example, \model{} takes only 10 minutes to model a dynamic object, {\it vs} 120 minutes by the previous art model Consistent4D.
Our code and video results are publicly available at \url{https://fudan-zvg.github.io/Efficient4D/}.
\end{abstract}

\section{Introduction}
\label{sec:intro}

Humans possess a remarkable capacity to comprehensively comprehend the spatial and temporal characteristics of a dynamic object in a brief video, even with a limited perspective, enabling them to predict its appearance in unseen viewpoints over time.
Despite the significant advancement of 3D object generation,
existing works \cite{poole2023dreamfusion, lin2022magic3d, chen2023fantasia3d, wang2023prolificdreamer} mostly consider static scenes or objects.
With the availability of large-scale 3D datasets~\cite{deitke2023objaverse, deitke2023objaversexl}, 
training generalizable models capable of directly generating multi-view images becomes possible \cite{liu2023zero, liu2023syncdreamer, long2023wonder3d}. These generated images can be turned into a 3D object through reconstruction techniques~\cite{mildenhall2021nerf, wang2021neus}. 
By further augmenting these generated static objects with predefined animations~\cite{maximo}, dynamic 3D content can be generated. However, this approach is still limited due to the need for fine-grained meshes as well as rigid restrictions. 

Directly generating 4D object/scene content from text description has been recently attempted \cite{singer2023text}. 
To bypass the need of exhaustively labeled training data pairs in form of (text, 4D), it trains a Neural Radiance Fields (NeRF)-like representation \cite{mildenhall2021nerf} via score distillation sampling \cite{poole2023dreamfusion} and separates the modeling of static scene and its dynamics. Not only is this method computationally inefficient caused by heavy supervision back-propagation through a large pretrained model, but also its textual condition is highly ambiguous in expressing the intended visual content. 
In quest of the aforementioned human's capability,
a recent work \cite{jiang2023consistent4d} proposed to generate dynamic 3D object images from a single-view video (statically captured monocular video from a fixed view), namely as {\it video-to-4D} object generation. However, similar as \cite{singer2023text} this method is also slow to train (\eg, 120 minutes to model a single dynamic object) in addition to complex design, hence unscalable and expensive in practice.

\begin{figure*}[t]
    \centering 
    \setlength{\tabcolsep}{0.0pt}
    \begin{tabular}{c}

    {\includegraphics[width=0.99\textwidth,clip]{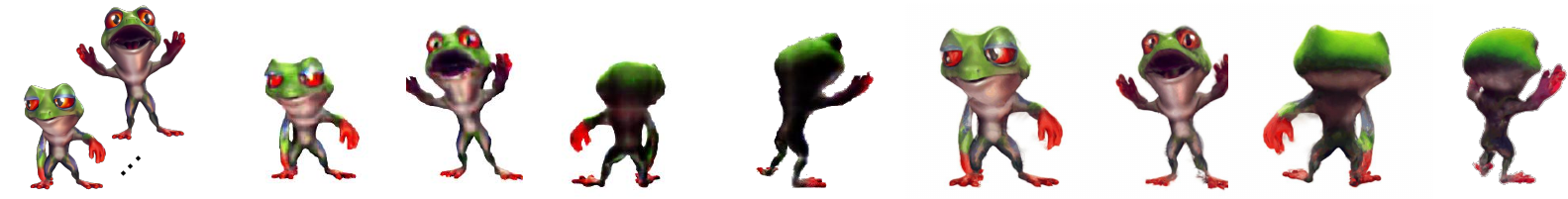}}
    \vspace{-2mm}
    \\
    \begin{tabularx}{1.0\linewidth}{ >{\centering\arraybackslash}p{0.15\linewidth} >{\centering\arraybackslash}X >{\centering\arraybackslash}X}
      \small{Input} &  \small{\makecell{Consistent4D~\cite{jiang2023consistent4d}: 120 mins}} & \small{\makecell{Efficient4D (Ours):  \textbf{10 mins}}}
    \end{tabularx}
    \\
    {\includegraphics[width=0.99\textwidth,clip]{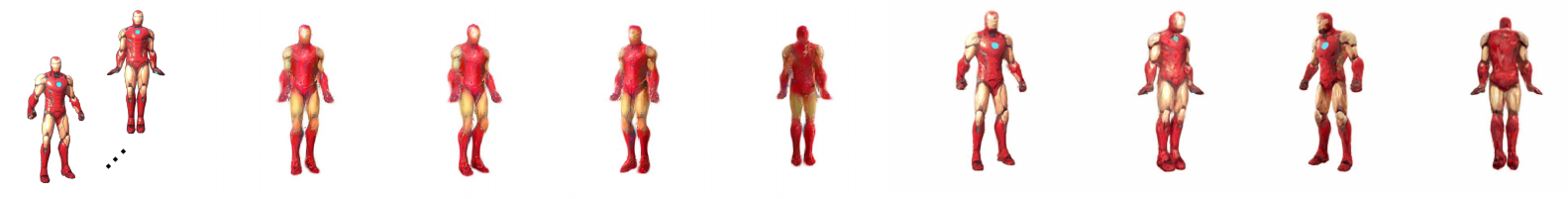}}
    \vspace{-2mm}
    \\
    \begin{tabularx}{1.0\linewidth}{ >{\centering\arraybackslash}p{0.15\linewidth} >{\centering\arraybackslash}X >{\centering\arraybackslash}X}
      \small{Input} &  \small{\makecell{4DGen~\cite{yin20234dgen}: 130 mins}} & \small{\makecell{Efficient4D (Ours): \textbf{10 mins}}}
    \end{tabularx}
    \\
    {\includegraphics[width=0.99\textwidth,clip]{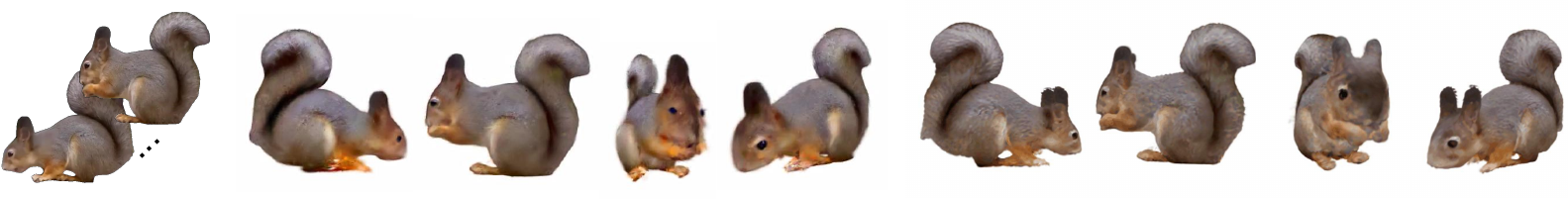}}
    \vspace{-2mm}
    \\
    \begin{tabularx}{1.0\linewidth}{ >{\centering\arraybackslash}p{0.15\linewidth} >{\centering\arraybackslash}X >{\centering\arraybackslash}X}
      \small{Input} &  \small{\makecell{STAG4D~\cite{zeng2024stag4d}: 70 mins}} & \small{\makecell{Efficient4D (Ours): \textbf{10 mins}}}
    \end{tabularx}
    
    \end{tabular}
    
    \caption{Examples of video-to-4D generation.
    {\bf Input}: A brief video of a dynamic object, as represented by 2 frames per case;
    {\bf Output}: Generated novel views at different timestamps. The generation time is also shown for each method. More visualized comparisons can be found in Figure~\ref{fig:quality_video}.
    }

\label{fig:teaser}
\end{figure*}

To address identified limitations, we formulate an efficient video-to-4D two staged object generation method called {\bf\it \model}.
In the first stage, we generate spacetime consistent images across different camera views as synthetic training data.
This is realized by imposing temporal smoothing into a multi-view image generator (\eg, SyncDreamer~\cite{liu2023syncdreamer}) in tandem with frame interpolation.
In the second stage, we use these training data to optimize a 4D Gaussian splatting model~\cite{yang2023gs4d}.
This is an extension of the 3D Gaussian splatting \cite{kerbl3Dgaussians}, originally designed for static 3D scene representation, with the temporal dimension introduced additionally, allowing for real-time rendering under continuous camera trajectories.
Using Gaussian representation brings about further computational efficiency gain, when compared with NeRF based designs (Figure~\ref{fig:teaser}).
To tackle the challenging discontinuity between the generated sparse frames, we design an inconsistency-aware loss function based on whether there is confidence in the consistency of a frame with its adjacent frames.
Along with a lightly weighted score distillation sampling loss for smooth viewpoint transitions, this technique enables robust 4D reconstruction. Notably, although our method takes a single-view video as input, \model{} can be easily extended to image-to-4D task by leveraging a image-to-video diffusion model~\cite{blattmann2023stable}.

Our {\bf contributions} are summarized as follows:
\textbf{(i)} 
We consider for the first time the efficiency challenge with the under-studied video-to-4D object generation problem.
\textbf{(ii)}
We propose an efficient video-to-4D object generation pipeline,
{\it \model{}}, characterized by directly generating high-quality training data without the need for heavy supervision back-propagation through a large pretrained model as suffered by most 3D/4D object generation approaches. We also extends \model{} to image-to-4D task.
\textbf{(iii)}
We introduce a inconsistency-aware confidence-weighted  loss for reconstructing a 4D Gaussian splatting model using the generated training data efficiently and robustly.
\textbf{(iv)}
Extensive experiments on both synthetic and real data validate the 
significant efficiency advantage (\eg, 10$\times$ speedup) of our \model{} over the prior art,
whilst maintaining the quality of novel view synthesis.
Also, our method can work well under the more challenging few-shot setting where only a handful of key frames are available for training, further extending the application scope.

\section{Related work}
\label{sec:related}

\noindent\textbf{3D generation}

3D generation takes two main settigns: text-to-3D and image-to-3D. 
The pioneering work, DreamFusion \cite{poole2023dreamfusion}, introduces the score distillation sampling (SDS) loss for optimizing 3D shapes with diffusion models. 
SDS's generality has prompted numerous subsequent efforts in both text-to-3D tasks \cite{lin2022magic3d, metzer2022latent, seo2023let, shi2023mvdream, chen2023fantasia3d, li2023focaldreamer, tsalicoglou2023textmesh, huang2023dreamtime, wu2023hd, wang2023prolificdreamer, zhu2023hifa} and image-to-3D tasks~\cite{xu2023neurallift360, tang2023make, melas2023realfusion, qian2023magic123} across various dimensions.
However, SDS-based approaches often suffer from difficulty to converge and extended optimization times. Conversely, some efficient methods~\cite{nichol2022point, shape, liu2023zero, liu2023one, liu2023syncdreamer} have emerged. Notably, Point-E~\cite{nichol2022point} and Shap-E~\cite{shape} train models to directly generate 3D point clouds or meshes. 
Zero123~\cite{liu2023zero} focuses on generating a 2D image from an unseen view based on a single image, convertible to a 3D shape through SDS or \cite{liu2023one}. 
Importantly, SyncDreamer~\cite{liu2023syncdreamer} produces multi-view consistent images, offering inspiration for reconstructing 3D objects.

\noindent\textbf{4D representation}

Efforts to synthesize videos with free-viewpoint control in dynamic scenes have a well-documented history~\cite{zitnick2004high}. 
For example, pre-NeRF~\cite{mildenhall2021nerf} approaches challenges in reconstructing intricate scene details. 
Recent advancements in 4D representations, particularly those based on neural rendering, include D-NeRF~\cite{pumarola2021dnerf}, DeVRF~\cite{liu2022devrf} and HyperNeRF~\cite{park2021hypernerf}, which decouple geometry and motion, utilizing a canonical space and a learned deformation field. 
DynIBaR~\cite{li2023dynibar} deploys an image-based rendering paradigm for representing long videos with complex camera and object motions. 
Another group of methods~\cite{fridovich2023kplanes,cao2023hexplane,shao2023tensor4d} adopt tensor decomposition of 4D volumes to represent dynamic 3D scenes.

Recently, Gaussian Splatting~\cite{kerbl3Dgaussians} has received widespread attention for its real-time high-fidelity rendering, especially its explicit point-based representation which holds great potential in modeling dynamic scenes. Consequently, a significant amount of work has been proposed to explore its extension to dynamic scene modeling. 
Among them, 
Deformable 3D Gaussians~\cite{yang2023deformable3dgaussian} and 4DGaussian~\cite{wu20234dgaussians} integrated the deformation field with 3D Gaussian Splatting for the joint learning of the scene's geometry and dynamics. SC-GS~\cite{huang2023sc} represents the motion with a set of sparse control points to achieve reconstruction and motion editing. Unlike the previous radiance field-based representations often involving complex training schedules or suffering from slow convergence, these methods can be optimized efficiently, achieving real-time rendering while surpassing the past methods in terms of quality. 

Our work is perpendicular to all the above works, where any of them can be deployed in our reconstruction phase. 
But considering both optimization efficiency and expressive capability, we choose to represent dynamic 3D assets by a set of native 4D scene primitives, which is proved to be superior in 4DGS~\cite{yang2023gs4d}. 

\noindent\textbf{4D generation}
There are a few recent works dedicated for more challenging 4D object generation.
For instance, MAV3D~\cite{singer2023text} deals with a text-to-4D problem by training Hexplane~\cite{cao2023hexplane} with a video diffusion model and SDS loss. 
Instead of text input, Consistent4D~\cite{jiang2023consistent4d} conditions the generation of 4D object over time on a monocular video with richer and more specific information.
However, it is computationally inefficient due to inheriting the previous SDS loss, along with complex pipeline design.
To overcome this limitation, we present a novel two-staged pipeline in a generation-and-reconstruction strategy, drastically boosting the training speed by 20$\times$ whilst maintaining the quality of novel view synthesis.

Recent works have attempted to generate 4D videos from a single input video. For example, TrajectoryCrafter \cite{yu2025trajectorycrafter} leverages rendered point cloud as the condition of diffusion model to facilitate novel view synthesis. However, the point cloud is projected from the source video and its estimated depth, so there are no points from the back view, limiting the range of novel views. 
Another work ReCamMaster~\cite{bai2025recammaster} trains a diffusion model which introduces frame dimension conditioning for input video and camera pose conditioning for target trajectory. But the trajectory is still limited to the front 180$^\circ$.
In contrast, our goal is to generate the complete 360$^\circ$ views for a dynamic object.


\section{Preliminary}
\label{sec:pre}

\noindent\textbf{4D Gaussian splatting}

The 4D Gaussian splatting (4DGS~\cite{yang2023gs4d}) builds upon the 3D Gaussian splatting technique introduced in~\cite{kerbl3Dgaussians}, originally designed for static scene representation. To address the complexities of dynamic scenes, 4DGS represents each Gaussian defined as:
\begin{equation}
\label{eq:gaussian_function}
G(\mathbf{p}|\mathbf{\mu}, \Sigma) = e^{-\frac{1}{2} \left( \mathbf{p}-\mathbf{\mu} \right)^\top \Sigma^{-1} \left( \mathbf{p}-\mathbf{\mu} \right)},
\end{equation}
where $\mathbf{\mu} \in \mathbb{R}^4$ is the mean vector, and $\Sigma \in \mathbb{R}^{4\times 4}$ is the anisotropic covariance matrix. The input $\mathbf{p} = (\mathbf{x}, t) \in \mathbb{R}^4$ represents a spacetime position with a spatial coordinate $\mathbf{x}$ and time $t$. The covariance matrix $\Sigma$ decomposes into a diagonal scaling matrix $S \in \mathbb{R}^{4\times 4}$ and a rotation matrix $R \in \mathbb{R}^{4\times 4}$ through $\Sigma = R S S^\top R^\top$. The 4D rotation $R$ is represented by a pair of iso-rotations, each characterized by a quaternion.

For rendering, each Gaussian includes opacity $\alpha$ and view-dependent color $\mathbf{c}$ represented by spherical harmonics (SH). Given an arbitrary view $\mathcal{I}$ defined by intrinsic and extrinsic parameters, we render the pixel at position $(u,v)$ with timestamp $t$ by blending visible Gaussians:
\begin{align}
\mathcal{I}(u,v,t) = &\sum_{i} f_{i}(u,v,t) \alpha_i \mathbf{c}_{i}(d)T_i
\\
\text{with }
T_i = &\prod^{i-1}_{j=1} (1- f_{j}(u,v,t) \alpha_j).
\end{align}
Here, $i$ indexes the visible Gaussians sorted by depth, $d$ refers to the direction of the pixel under the view $\mathcal{I}$, and $f_{i}(u,v,t)$ denotes the influence of a Gaussian on this position. To obtain the influence, unlike 3D Gaussian Splatting~\cite{kerbl3Dgaussians} which directly projects 3D Gaussians to image space, we need condition 4D Gaussians on time and then project them. More specifically, $f(u,v,t)$ is expressed by:
\begin{equation}
f(u,v,t) = G_t(t|\mathbf{\mu}_t, \Sigma_t)G_{{P}(\mathbf{x})|t}(u,v|{P}(\mathbf{\mu}_{\mathbf{x}|t}), {P}(\Sigma_{\mathbf{x}|t})),
\end{equation}
where $G_t$ is the marginal distribution of the 4D Gaussian in time, and $G_{\text{P}(\mathbf{x})|t}$ is the projected version of the conditional 3D Gaussian with
\begin{align}
    \mathbf{\mu}_{\mathbf{x}|t} &= \mathbf{\mu}_{\mathbf{x}} + \Sigma_{\mathbf{x},t}\Sigma_{t}^{-1} (t - \mathbf{\mu}_t),\\
    \Sigma_{\mathbf{x}|t} &= \Sigma_{\mathbf{x}} - \Sigma_{\mathbf{x},t}\Sigma_{t}^{-1}\Sigma_{t,\mathbf{x}}.
\end{align}
The projection operation $P$~\cite{kerbl3Dgaussians, Zwicker2001EWA} projects the world point $\mathbf{\mu}_{\mathbf{x}|t}$ to image space and transforms the covariance by 
\begin{equation}
   {P}(\Sigma_{\mathbf{x}|t}) = JW\Sigma_{\mathbf{x}|t}W^\top J^\top 
\end{equation}
where $W$ is the extrinsic matrix of $\mathcal{I}$ and $J$ is the Jacobian of the affine approximation of the projective transformation.

\noindent\textbf{Score distillation sampling}
Score distillation sampling (SDS) is first introduced by DreamFusion~\cite{poole2023dreamfusion}, which is used to distill the knowledge from a pretrained diffusion model $\epsilon_{\phi}$. Specifically, given an image $I$ rendered from a scene representation (\eg 3DGS) parameterized by $\theta$, the gradient of SDS loss is calculated as:
\begin{equation}
    \nabla_{\theta} \mathcal{L}_{\mathrm{SDS}} (\phi, I_t) = \mathbb{E}\left[ w(t) \left(\epsilon_{\phi}(I_t; t, c) - \epsilon \right) \frac{\partial I_t}{\partial \theta} \right],
\end{equation}
where $I_t$ is the perturbed image with noise $\epsilon$ at time step $t$, and $c$ is the condition (\eg one frame of the input video in this paper).

\section{Method}
\label{sec:method}

\begin{figure*}[t]
    \centering 
    \setlength{\tabcolsep}{0.0pt}
    \begin{tabular}{c} 

    {\includegraphics[width=1.0\textwidth,clip]{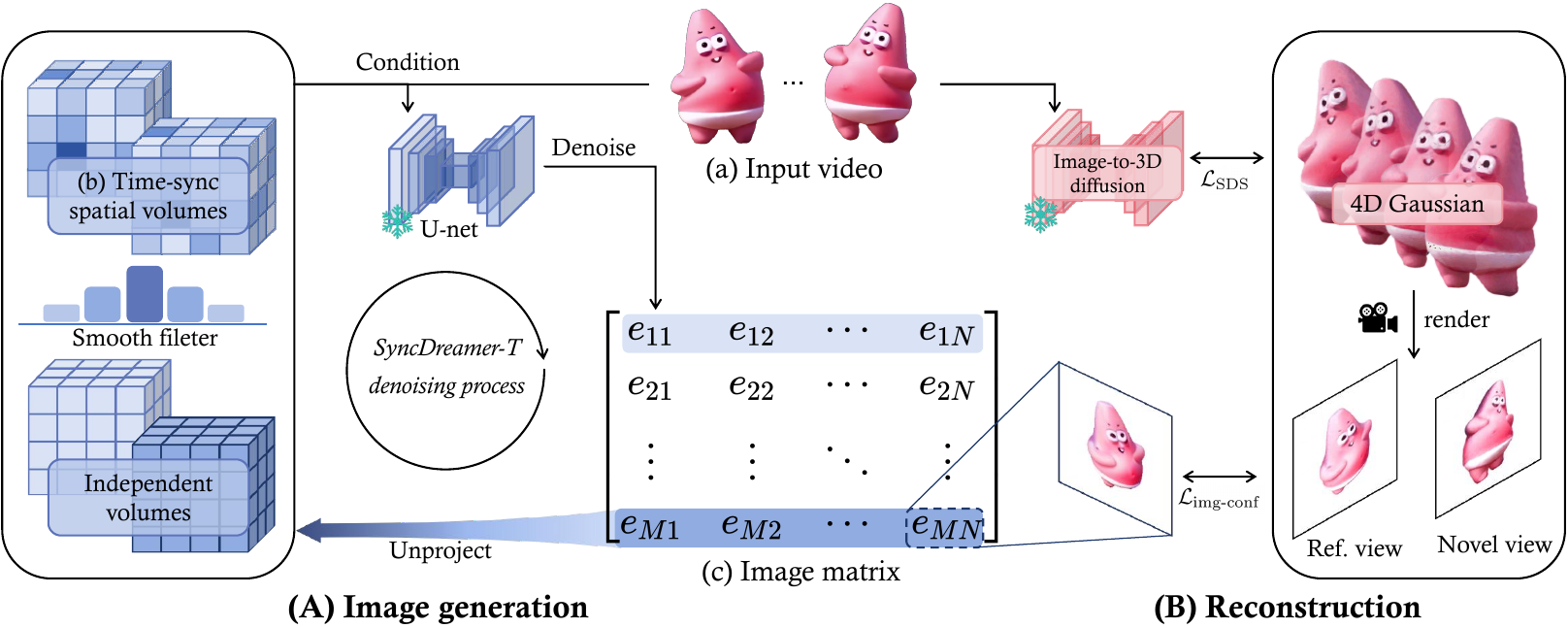}}
    \\
        
    \end{tabular}
    \caption{\textbf{Overview of our \model{} approach.} 
    Given as the input (a) a brief video depicting a dynamic object from a single perspectives,
    our model aims to generate this object with geometrical and temporal consistency under any specific view and time.
    \model{} comprises two components: 
    {\bf(A)} Image sequence synthesis through (b) time-synchronous spatial volumes, resulting in (c) an {\it image matrix}
    where each row consists of multi-view geometrically consistent images
    and each column consists of view-specific temporally consistent images.
    {\bf(B)} 4D Reconstruction using the generated images in (A). The 4D Gaussian representation can be trained efficiently and robustly under the confidence-weighted loss $\mathcal{L}_{\text{img-conf}}$ and the low-weighted SDS loss $\mathcal{L}_{SDS}$.
    }
    \label{fig:pipeline}

\end{figure*}

Our \model{} addresses the challenge of efficiently generating dynamic objects under novel views from a single-view video. The input single-view video can be either provided by user or generated by a video generation model. The latter approach extends the application of our method beyond video-to-4D. For example, we can also achieve image-to-4D through a image-to-video diffusion model~\cite{blattmann2023stable}.
As illustrated in Figure~\ref{fig:pipeline}, it comprises two key components:
\begin{itemize}
    \item An image  synthesis pipeline (Figure~\ref{fig:pipeline}(A)) generates images across views and timestamps, ensuring sufficient geometry and temporal consistency.

    \item An efficient and robust reconstruction process (Figure~\ref{fig:pipeline}(B)) efficiently utilizes the synthetic images 
    for accurate dynamic object reconstruction and novel view synthesis.
\end{itemize}

We will elaborate on these components in Section~\ref{sec:image_gen} and \ref{sec:reconstruction}, respectively.

\subsection{Image synthesis across views and timestamps}
\label{sec:image_gen}

Due to the difficulty of obtaining calibrated 4D scans, our approach involves the direct generation of high-quality consistent 4D data from a single-view video which is much easier to capture (Figure~\ref{fig:pipeline}(a)). Specifically, we seek to produce a $M\times N$ {\bf\it image matrix} $\mathcal{D} = \{e_{ij}\}_{i,j=1}^{M,N}$ representing 2D images with geometrical and temporal consistency. Here, $M$ denotes timestamps, and $N$ represents views, with each matrix element corresponding to an image (Figure~\ref{fig:pipeline}(c)). This approach combines conventional video (capturing time variation, represented by a single column in the image matrix) and 3D (capturing view variation, represented by a single row in the image matrix) generation \cite{singer2022make, liu2023syncdreamer}, offering comprehensive information for modeling a dynamic object.

To initiate the image matrix $\mathcal{D}$, we set the first view (\ie, the first column) with $K$ frames from the input video and proceed to generate the remaining views. Our task involves generating multi-view consistent images from a single image for each row. Existing image-to-3D methods, such as SyncDreamer~\cite{liu2023syncdreamer}, can be leveraged for this purpose. However, these methods often struggle with temporal inconsistency within a specific view (\ie, continuity in the column direction) due to the independent synthesis of multi-frame images.
To address this issue, we propose an enhanced version of SyncDreamer with improved temporal continuity, referred to as {\bf\it SyncDreamer-T}.

Specifically, SyncDreamer generates $N$ multi-view images $\{\mathbf{x}_0^{(1)}, \dots, \mathbf{x}_0^{(N)}\}$ of a static object using a synchronized $N$-view noise predictor $\{\epsilon_{\theta}^{(n)} | n=1,\dots, N\}$ that predicts synchronized noise for noisy multi-view images $\mathbf{x}_t^{1:N}$. The noise predictor is conditioned on information correlated with all views.
Cross-view conditioning is achieved through a spatial feature volume $\mathcal{V} \in \mathbb{R}^{F\times V\times V\times V}$ unprojected by $\mathbf{x}_t^{1:N}$ to inject 3D-aware features into the shared noise predictor, ensuring geometrical consistency across views for static moments.

To impose temporal consistency, the information from $\mathcal{V}$ at different timestamps is aggregated using a time-synchronous spatial volume we design here (Figure~\ref{fig:pipeline}(b)). A smoothing filter layer is introduced into the spatial volumes of different frames/timestamps, incorporating a weight vector $\mathbf{w} = (w_{-k}, \dots, w_0, \dots, w_k)$ which serves as the smooth filter. At each denoising step, time-synchronized spatial volumes $\Tilde{\mathcal{V}}_i$ for each input frame $i \in \{1, 2, \dots, M\}$ are constructed as:
\begin{equation}
\label{eq:smooth_v}
    \Tilde{\mathcal{V}}_i = \sum_{j=-k}^{k} w_{i+j} \mathcal{V}_{i+j}.
\end{equation}
This synchronization ensures consistent features across 
$k$ past and $k$ future frames
during the denoising process, thus enhancing temporal consistency. With this time-synchronized spatial volumes, the proposed {\bf\it SyncDreamer-T} is entirely {\it training-free} established on the pretrained SyncDreamer.

To further improve temporal resolution, video frame interpolation (\eg, RIFE~\cite{huang2022real}) can be applied after generating the image matrix $\mathcal{D}$. 
The midpoint interpolation is applied twice recursively, giving three additional frames between each pair of consecutive frames. This results in a total of $M = 4K-3$ images in a column of $\mathcal{D}$. 

\begin{figure*}[t]
    \centering 
    \setlength{\tabcolsep}{0.0pt}
    \begin{tabular}{c:c} 

    \raisebox{.12\height}
    {\includegraphics[width=0.1\textwidth,clip]{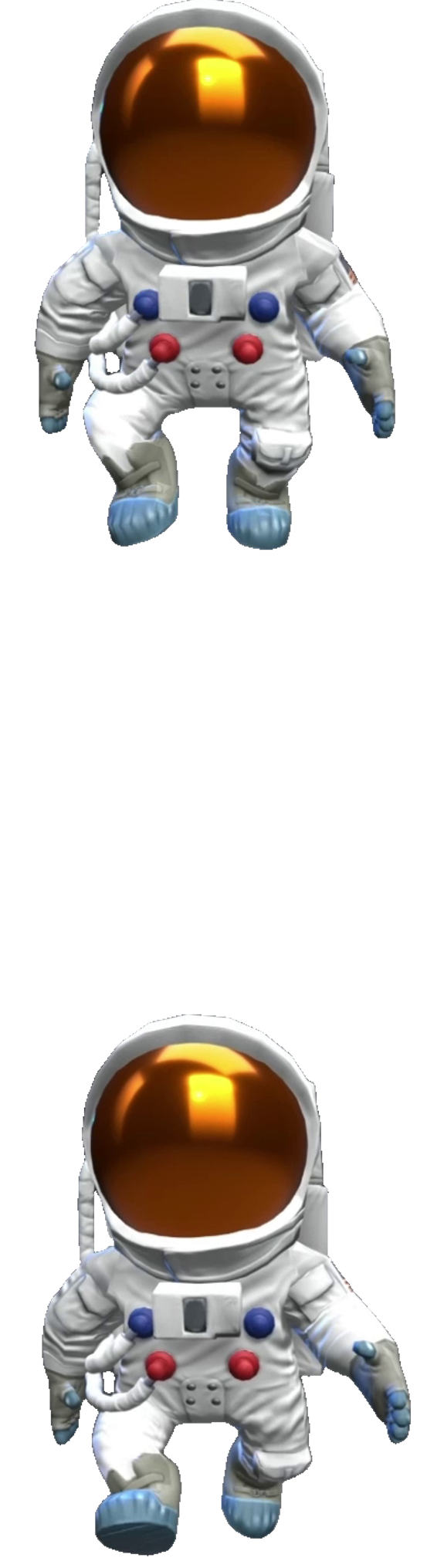}}~
    &
    {\includegraphics[width=0.8\textwidth,clip]{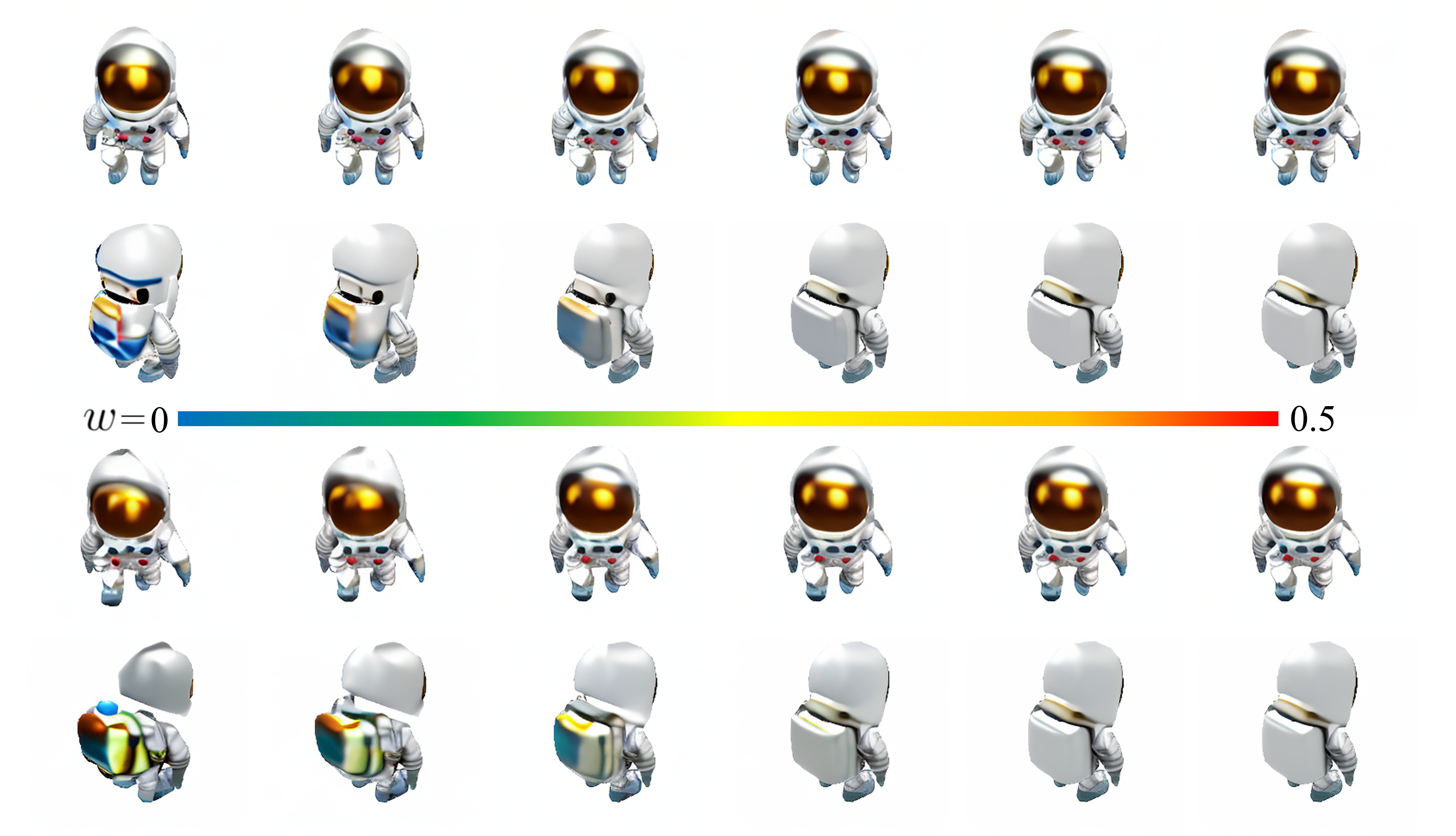}}

    \\
    Input & Feature fusion

    \end{tabular}
    \caption{
    {\bf Example analysis on temporal synchronization}.
    In this illustration, we manipulate the fusion ratio ($w$) across a range from 0 to 0.5 for the spatial feature volumes of two input frames. The results indicate that a moderate ratio can achieve superior outcomes by balancing both temporal consistency and motion independence.
    }
\label{fig:time_sync}
\end{figure*}
\noindent\textbf{Analysis on temporal synchronization}
For a clearer insight into our temporal synchronization design, we undertake a simplified experiment involving two input frames, each with its feature volumes labeled as $\mathcal{V}_1$ and $\mathcal{V}_2$. The fusion process is carried out by combining them as follows:
\begin{align}
    \Tilde{\mathcal{V}}_i = (1 - w) \mathcal{V}_i + w \mathcal{V}_{3-i},
    \label{eq:fuse_feature_2}
\end{align}
where $i\in \{1,2\}$ and $w$ denotes the fusion ratio, we systematically vary the ratio $w$ from 0 to 0.5, resulting in distinct columns of ``Feature fusion" as illustrated in Figure~\ref{fig:time_sync}. For $w=0$, the original generation process is represented, where the spatial volumes of the two frames are independent, leading to temporal inconsistency. As $w$ approaches $0.5$, we observe a gradual convergence of the generated astronauts conditioned on different input images, achieving similarity in both texture and geometry. This suggests that smoothing at the feature level is effective in aligning frames over time.

However, a challenge arises as the fused features may induce similar motion. While the bottom-left astronaut is stepping forward, the one at the bottom-right column does not exhibit forward motion like the top astronaut. Thus, a trade-off is necessary between achieving temporal consistency and preserving motion independence. In practical terms, it is recommended to set the ratio just above $0.5$, striking a balance that ensures temporal texture consistency at a moderate cost of entangled geometry. Also note that the results in Figure~\ref{fig:time_sync} are not sensitive to the choice.

\subsection{4D generation through reconstruction}
\label{sec:reconstruction}

Aiming for 3D dynamic object modeling, 
discrete images not suffice. 
Our next goal is to model a truly 4D content from the image matrix $\mathcal{D}$. 
For efficient modeling,
we have formulated the 4DGS representation model in Section~\ref{sec:pre},
departing from previous slow-to-train 4D reconstruction models~\cite{fridovich2023kplanes}. 

\noindent\textbf{Optimization}
In the training of 4DGS, optimization is performed on the mean ($\mathbf{\mu}$), covariance ($\Sigma$), opacity ($\alpha$), and spherical harmonic (SH) coefficients, as well as density control including densification and pruning for each Gaussian. 
The original loss function, as presented in~\cite{yang2023gs4d}, involves both RGB loss and SSIM loss with balancing weights fixed. Specifically, the loss formulation is defined as:
\begin{equation}
    \mathcal{L}_{\text{img}} = \lambda_{\text{rgb}} \mathcal{L}_{\text{rgb}} + \lambda_{\text{ssim}} \mathcal{L}_{\text{ssim}},
    \label{eq:loss}
\end{equation}
where $\mathcal{L}_{\text{rgb}}$ is the $L_1$ loss in RGB space, $\mathcal{L}_{\text{ssim}}$ is the SSIM loss~\cite{wang2004image}, and $\lambda_{\text{rgb/ssim}}$ is the respective weight hyper-parameter.

However, such optimization approach assumes clean training data, which may not valid for synthetic data with inherent imperfections. To address this, we first introduce a inconsistency-aware loss formulation with adaptive balancing weights:
\begin{equation}
    \lambda_{\text{rgb}} = \Tilde{\lambda}_{\text{rgb}} \mathcal{C}_{\text{rgb}},\quad
    \lambda_{\text{ssim}} = \Tilde{\lambda}_{\text{ssim}} \mathcal{C}_{\text{ssim}},
\end{equation}
where $\Tilde{\lambda}_{\text{rgb/ssim}}$ is fixed weight and
$\mathcal{C}_{\text{rgb/ssim}}$ is the adaptive confidence score of a generated image $\mathbf{I}$ calculated as
\begin{equation}
    \mathcal{C}_{\text{rgb}} = 1 - |\mathbf{I} - \hat{\mathbf{I}}|, \quad
    \mathcal{C}_{\text{ssim}} = \text{SSIM}(\mathbf{I}, \hat{\mathbf{I}}) ,
    \label{eq:conf}
\end{equation}
where $\hat{\mathbf{I}}$ is the unwarped image from adjacent frames estimated by optical flow. In such way, the confidence $\mathcal{C}_{\text{rgb/ssim}}$ function as an adaptive role in controlling the loss and gradient by assigning lower weights to inconsistent regions, thus enhancing overall reconstruction quality. 

The confidence design can guarantee temporal consistency, but achieving high quality of novel view synthesis still remains challenging due to the sparsity of the generated images. Therefore, we also incorporate SDS loss with a relatively small weight $\lambda_{\text{SDS}}$ for smooth transition across different supervised views. We use the image-to-3D diffusion model~\cite{blattmann2023stable} in SDS loss conditioned on the frame of input video at each timestamp. The total loss function is expressed by:
\begin{equation}
\label{eq:loss_total}
    \mathcal{L}_\text{total} = \mathcal{L}_\text{img-conf} + \lambda_{\text{SDS}}\mathcal{L}_{\text{SDS}}
\end{equation}
with $\mathcal{L}_\text{img-conf} = \Tilde{\lambda}_{\text{rgb}} \mathcal{C}_{\text{rgb}} \mathcal{L}_{\text{rgb}} + \Tilde{\lambda}_{\text{ssim}} \mathcal{C}_{\text{ssim}} \mathcal{L}_{\text{ssim}}$.

\section{Experiments}
\label{sec:exp}

\subsection{Experiment setup}
\label{sec:imple_detail}

\noindent\textbf{\textbf{Competitors}}
For comparison, we mainly focus on video-to-4D and image-to-4D task. The competitors include Consistent4D~\cite{jiang2023consistent4d} (ICLR2024), 4DGen~\cite{yin20234dgen} (ArXiv2024) and STAG4D~\cite{zeng2024stag4d} (ECCV2024) for video-to-4D, and Animate124~\cite{zhao2023animate124} (ArXiv2023) and DreamGaussian4D~\cite{ren2023dreamgaussian4d} (ArXiv2024) for image-to-4D.
We obtained their results by running their released official code. We also compare SyncDreamer~\cite{liu2023syncdreamer} partially by replacing SyncDreamer-T in ablation study (Section~\ref{sec:ablation}).

\noindent\textbf{\textbf{Evaluation data}}
To showcase the versatility of our proposed method, we conducted extensive experiments using a diverse set of data sources. 
For video-to-4D, we focused on 36 sequences: 32 sequences released by \cite{jiang2023consistent4d} and 4 sequences processed by ourselves. Among the released data, seven sequences are synthetic data where ground truth are available. For image-to-4D, we used the ten images released by \cite{zhao2023animate124}.
Our four sequences are used for sparse input evaluation in Section~\ref{sec:sparse}, which only contain two frames for each sequence. Three of them, named \textit{dragon}, \textit{guard}, and \textit{yoxi}, are rendered from 3D animated models obtained from Sketchfab~\cite{sketchfab}. The other one named \textit{yellow face} was collected from the internet. All the data are publicly available.

\noindent\textbf{\textbf{Evaluation metrics}}
\label{sec:metric}
As 4D generation research is still at early stage, 
there is no well established metric yet. However, we adopt multiple metrics by referring related works for comprehensive evaluation. For evaluating on synthetic data, we use LPIPS score~\cite{zhang2018unreasonable} and CLIP similarity~\cite{clip} between rendered images and ground truth following \cite{jiang2023consistent4d}. 
For the cases without ground truth, we also use CLIP-similarity between generated images and input frames as \cite{tang2023dreamgaussian, ren2023dreamgaussian4d} to measure image quality. For temporal smoothness, following \cite{yin20234dgen, esser2023structure, geyer2023tokenflow} we use CLIP-T to measure the similarity between adjacent frames of a generated video from different views, including front (CLIP-T-f), side (CLIP-T-s) and back (CLIP-T-b) views. 
To evaluate a 4D object completely for different methods, we obtain the generated images by rendering 320 images uniformly distributed in space and time, covering 16 viewpoints and 20 timestamps.

\noindent\textbf{\textbf{Implementation details}}
In equation~(\ref{eq:smooth_v}), the spatial volumes are smoothed locally in a sliding window style, so the consistency of generated images may be weakened as the frame number increase. In practice, we can adjust the weight following two rules: (i) the middle weight should be at least $0.5$; (ii) At least half of the volumes should have a positive weight. Here we provide some sample weights which have a good effect:
\begin{equation}
    \mathbf{w} = 
    \left\{\begin{matrix} 
        (1.5, 7, 1.5), \quad 3 \le N < 5 \\
        (1, 1, 6, 1, 1), \quad 5 \le N < 7 \\
        (1, 1, 1, 6, 1, 1, 1), \quad 7 \le N < 13 \\
        (1, 1, 1, 1, 1, 2, 14, 2, 1, 1, 1, 1, 1), \quad N \ge 13,
    \end{matrix}\right. 
\end{equation}
where $N$ denotes the frame number. The weight will be normalized to ensure the sum is $1$.

In denosing process, we follow the default setting of \cite{liu2023syncdreamer} using a improved sampling strategy introduced by HarmonyView~\cite{woo2023harmonyview}. HarmonyView redefines the score estimation by decomposing consistency and diversity. Please refer to \cite{woo2023harmonyview} for details.

In the reconstruction stage, Gaussians are initialized randomly inside the sphere with radius space 0.5 with identity rotations and a initial number of 50,000 without densification. Training of 4D Gaussian Splatting is carried out using the Adam optimizer for 500 iterations with batch size 1. All other 4DGS hyperparameters remain consistent with those in \cite{yang2023gs4d}. The balancing weights in loss function (equation~\ref{eq:loss_total}) are set as $\Tilde{\lambda}_{\text{rgb}} = 8000, \Tilde{\lambda}_{\text{ssim}} = 2000, \lambda_{\text{SDS}} = 1$.
In equation~(\ref{eq:conf}), the calculation of estimated image $\mathbf{I}$ is implemented by frame interpolation using a optical flow based model~\cite{huang2022real}. For each generated frame $I_t$, we use its adjacent four frames to interpolate the estimated frame as follows:
\begin{equation}
    \hat{I}_{t1} = \text{Interp}(I_{t-1}, I_{t+1}),~~
    \hat{I}_{t2} = \text{Interp}(I_{t-2}, I_{t+2}).
\end{equation}
Then we can compute the mean RGB confidence score:
\begin{equation}
    \mathcal{C}_{\text{rgb}} = 1 - \frac{1}{2} \left(|I_t - \hat{I}_{t1}| + |I_t - \hat{I}_{t2}|\right). 
\end{equation}
The SSIM confidence score is similar by using equation~(\ref{eq:conf}).

For speed efficiency, our proposed method only costs about 2 minutes for image generation if parallel denoising is allowed and 8 minutes for reconstruction on one A6000 GPU.

\subsection{Evaluation on synthetic data}
\begin{figure}[tb]
  \centering
    \setlength{\tabcolsep}{0.0pt}
    \begin{tabular}{c} 

    {\includegraphics[width=0.99\linewidth,clip]{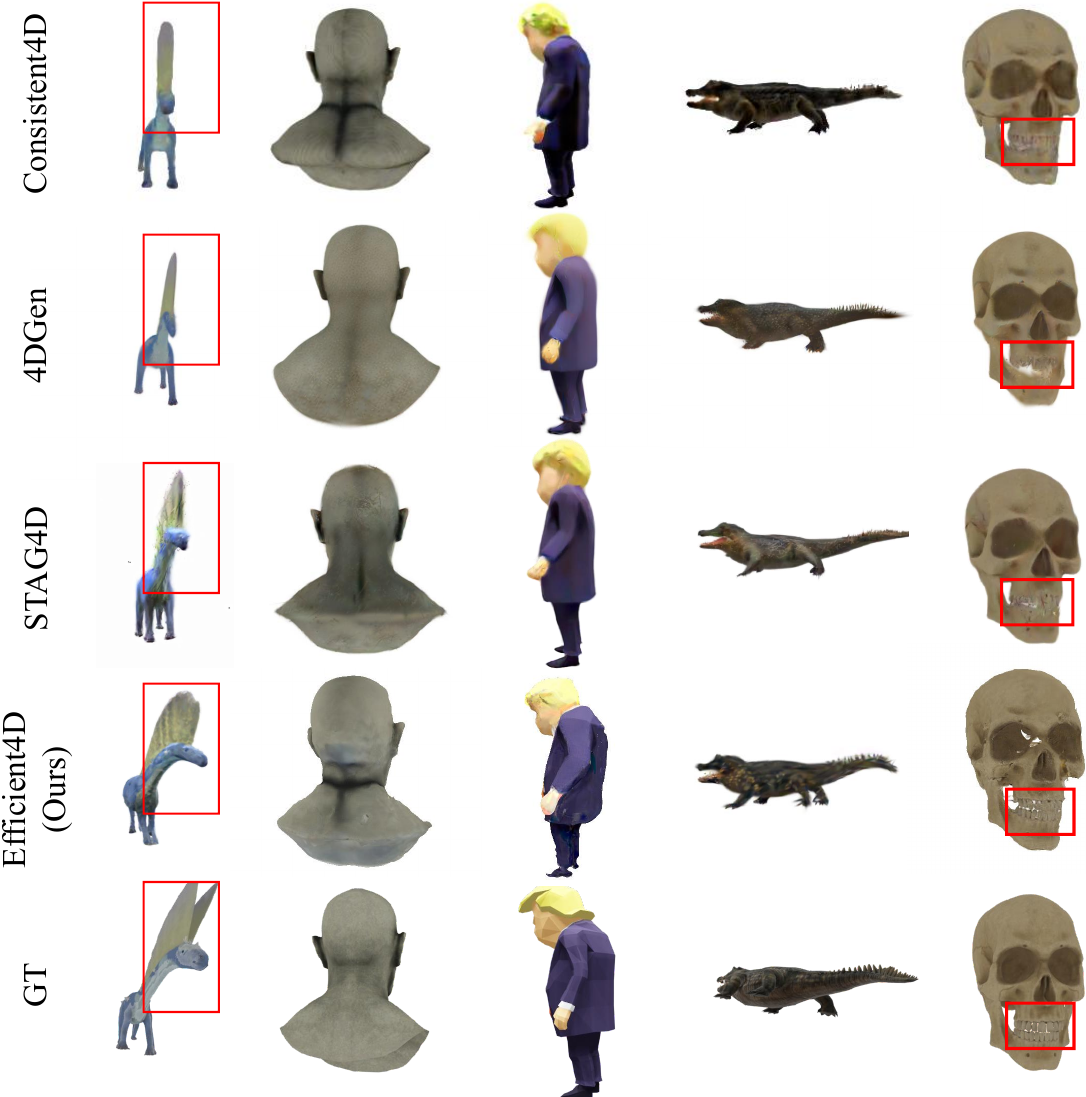}}
    
    \end{tabular}
    \caption{\textbf{Qualitative evaluation} against ground truth (GT) on synthetic data. We compare our \model{} with Consistent4D~\cite{jiang2023consistent4d}, 4DGen~\cite{yin20234dgen} and STAG4D~\cite{zeng2024stag4d}. }
    \label{fig:qual_synt}

\end{figure}
\begin{table}[t]
\caption{\textbf{Quantitative evaluation} on synthetic data. We report CLIP and LPIPS scores between rendered images and ground truth images.}
\label{tab:quan_synt}

\begin{center}

 \setlength{\tabcolsep}{0.6mm}{

\begin{tabular}{l|cccc}

\hline

\hline

\hline

\hline

 & Consistent4D & 4DGen & STAG4D & \model (Ours) \\

\hline

\hline

CLIP $\uparrow$ & 0.87 & 0.89 & 0.91 & \bf 0.92 \\

 LPIPS $\downarrow$ & 0.16 & 0.14 & \bf 0.13 & \bf 0.13 \\

\hline

\hline

\hline

\hline

\end{tabular}
}

\end{center}

\end{table}

We first present the results on synthetic data in Figure~\ref{fig:qual_synt} with ground truth shown. Since the data only include videos, we compare with two video-to-4D methods: Consistent4D~\cite{jiang2023consistent4d}, 4DGen~\cite{yin20234dgen} and STAG4D~\cite{zeng2024stag4d}.
Our visual results exhibit superior accuracy in both texture and geometry when compared to the ground truth, such as the direction of dinosaur's head, the color of trump's arm and the teeth of skull. The quantitative metrics are also presented in Table~\ref{tab:quan_synt}. The better CLIP and LPIPS scores further validate our superior generation results.

\subsection{Video-to-4D comparison}
\begin{figure*}[t]
    \centering 
    \setlength{\tabcolsep}{0.0pt}
    \begin{tabular}{c} 

    {\includegraphics[width=1.0\textwidth,clip]{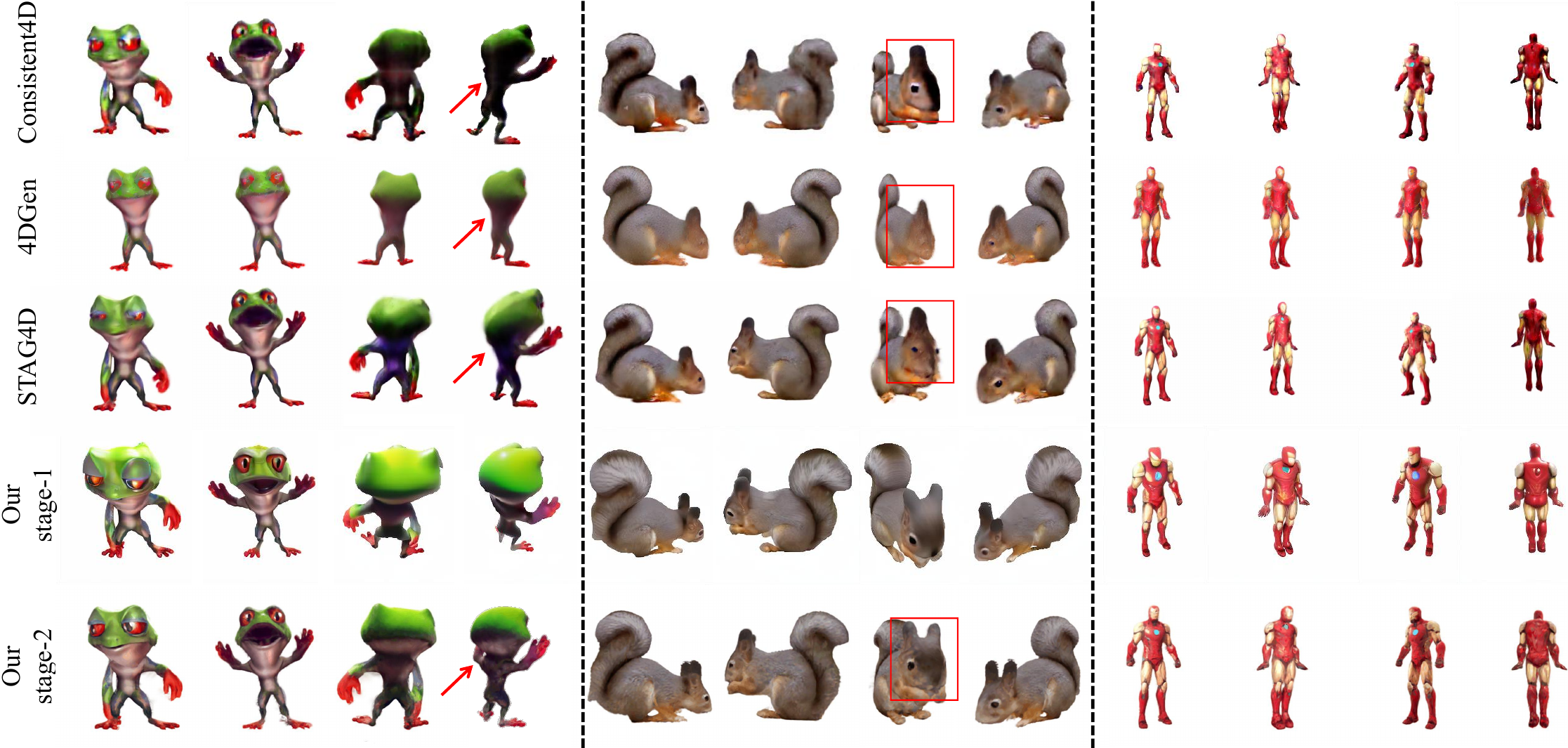}}

    \end{tabular}
    \caption{\textbf{Qualitative comparisons on video-to-4D generation.} We compare our \model{} with Consistent4D~\cite{jiang2023consistent4d}, 4DGen~\cite{yin20234dgen} and STAG4D~\cite{zeng2024stag4d}. For each case, we show four images per method with $0^{\circ}$ elevation. Our \model{} comprises two stages: image generation stage (Our stage-1, $30^{\circ}$ elevation) and reconstruction stage (Our stage-2, $0^{\circ}$ elevation).}
\label{fig:quality_video}
\end{figure*}
\begin{figure*}[t]
    \centering 
    \setlength{\tabcolsep}{0pt}
    \begin{tabular}{c} 

    {\includegraphics[width=1.0\textwidth,clip]{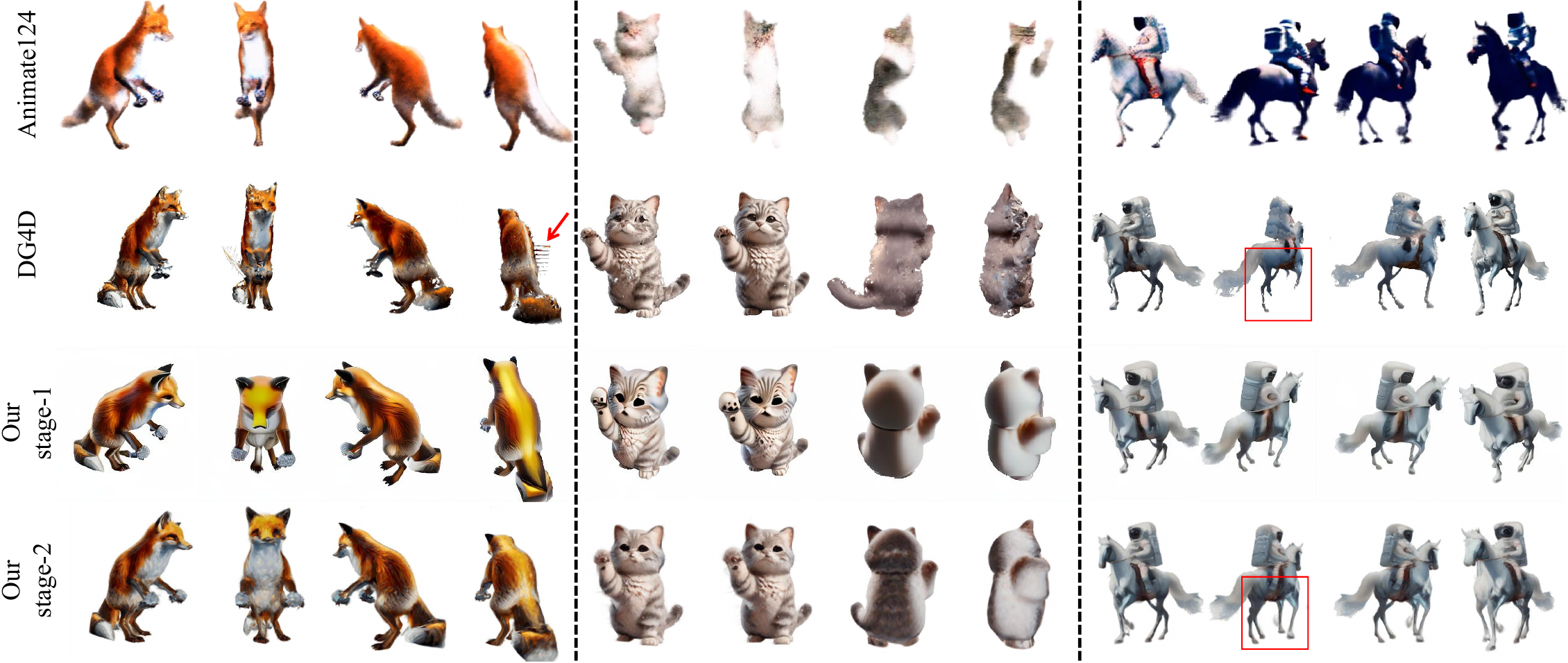}}

    \end{tabular}
    \caption{\textbf{Qualitative comparisons on image-to-4D generation.} We compare our \model{} with Animate124~\cite{zhao2023animate124} and DreamGaussian4D~\cite{ren2023dreamgaussian4d} (DG4D). For each case, we show four images per method with $0^{\circ}$ elevation. Our \model{} comprises two stages: image generation stage (Our stage-1, $30^{\circ}$ elevation) and reconstruction stage (Our stage-2, $0^{\circ}$ elevation).}
\label{fig:quality_image}
\end{figure*}
\begin{table*}[t]
\caption{\textbf{Quantitative comparisons} with state-of-the-art methods on both video-to-4D and image-to-4D generation.}
\label{tab:quan}

\begin{center}

 \setlength{\tabcolsep}{2.0mm}{

\begin{tabular}{l|ccccc}

\hline

\hline

\hline

\hline

Method & CLIP$\uparrow$ &  CLIP-T-f$\uparrow$ & CLIP-T-s$\uparrow$ & CLIP-T-b$\uparrow$ & Generation time$\downarrow$ \\

\hline

\hline

\multicolumn{6}{l}{\textit{- Video-to-4D comparison}}             \\

\hline

\hline

Consistent4D~\cite{jiang2023consistent4d} & 0.8471 & 0.9692 & \underline{0.9658} & 0.9697 & 120 mins \\

4DGen~\cite{yin20234dgen} & \underline{0.8730} & 0.9568 & 0.9568 & 0.9573 & 130 mins \\

STAG4D~\cite{zeng2024stag4d} & 0.8398 & \bf 0.9766 & \bf 0.9731 & \underline{0.9760} & 70 mins \\

\model{} (Ours) & \bf 0.8745 & \bf 0.9766 & 0.9609 & \bf 0.9780 & \bf 10 mins \\

\hline

\hline

\multicolumn{6}{l}{\textit{- Image-to-4D comparison}}             \\

\hline

\hline

Animate124~\cite{zhao2023animate124} & 0.8076 & \bf 0.9673 & \bf 0.9639 & \bf 0.9541 & 540 mins \\

DreamGaussian4D~\cite{ren2023dreamgaussian4d} & \underline{0.8242} & 0.8999 & 0.9072 & 0.9038 & 12 mins \\

\model{} (Ours) & \bf 0.8350 & \underline{0.9346} & \underline{0.9297} & \underline{0.9321} & \bf 10 mins \\

\hline

\hline

\hline

\hline

\end{tabular}
}

\end{center}

\end{table*}

For data without ground truth, we present qualitative comparisons in Figure~\ref{fig:quality_video}. 
When assessing texture quality, it is important to note that all methods fall under the category of lifting 2D to 4D. However, Consistent4D produce watercolor-like images with low fidelity, which can be exemplified by the blurry edges of their rendered images.
We attribute this blur to conflicts from multiple supervisory signals during prolonged optimization. In contrast, our method excels in directly generating high-quality 2D images, thus providing a strong and consistent supervision in the reconstruction stage. Although 4DGen and STAG4D also use pseudo labels, their images are inconsistent in the temporal coordinate and SDS loss still dominates the optimization. Therefore, 4DGen lacks details and STAG4D suffers from floaters and blur.
Besides, our method's reconstruction stage supports rendering under any viewpoints while the images generated in stage-1 are discrete and sparse with a fixed elevation $30^\circ$. Overall, our method consistently outperforms the baseline methods in most cases.

In Table~\ref{tab:quan}, we compare our method with baselines quantitatively. The metrics and data used are described in Section~\ref{sec:imple_detail}.
We draw several observations:
(i) Our method achieves superior image quality and temporal smoothness on all cases validated by the higher CLIP and CLIP-T scores;
(ii) Our method significantly accelerates the generation process, achieving over 10$\times$ speed improvement (10 mins {\it vs} 120 mins).
More specifically, the speed improvement is attributed to (i) our image supervision design, converging optimization faster and reduce the training iterations, and (ii) each iteration in our method requiring much less time thanks to the efficiency of Gaussian representation. 

\subsection{Image-to-4D comparison}

Next, we will evaluate our method on the image-to-4D task. In Figure~\ref{fig:quality_image}, we present the visual results of three different methods. It can be observed that our extended method is capable of generating 4D assets with higher quality compared to those produced by state-of-the-art image-to-4D methods. Animate124 generates textures lacking in detail, and DreamGaussian4D tends to produce fragmented meshes. In contrast, the results of our \model{} exhibit both good geometry and high-quality textures. Additionally, Table~\ref{tab:quan} supports the similar conclusion that our method is superior, as seen in the video-to-4D comparison. 
Although Animate124 achieves the best CLIP-T score, This score merely indicates that the range of motion is small rather than the motion being continuous. From Figure~\ref{fig:quality_image} we can easily find that the quality of Animate124 results are lower than any other methods.
Another notable point is that the overall CLIP scores for image-to-4D methods are lower than those for video-to-4D methods. This may be because the reference single-view videos in image-to-4D are generated by video diffusion models, which are inconsistent and have lower quality compared to real-world videos.

\subsection{Challenging cases}
\label{sec:sparse}

\begin{figure}[t]
    \centering 
    \setlength{\tabcolsep}{0.0pt}
    \begin{tabular}{c:c:c} 

    \raisebox{.1\height}
    {\includegraphics[width=0.055\textwidth,clip]{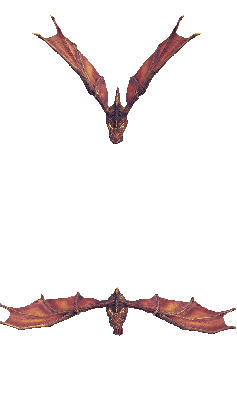}}~
    &
    \raisebox{.15\height}
    {\includegraphics[width=0.2\textwidth,clip]{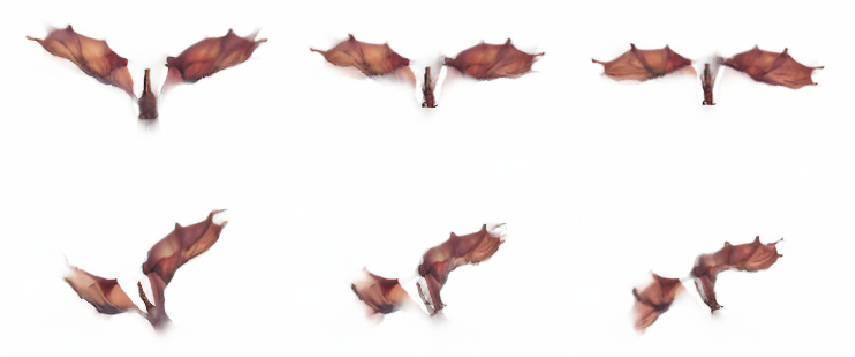}}~
    &
    \raisebox{.\height}
    {\includegraphics[width=0.19\textwidth,clip]{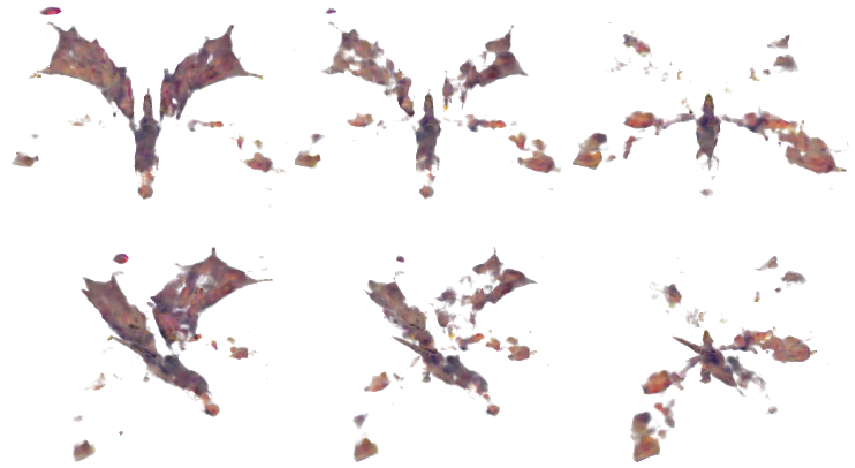}}
    
    \\
    \raisebox{-.01\height}
    {\includegraphics[width=0.057\textwidth,clip]{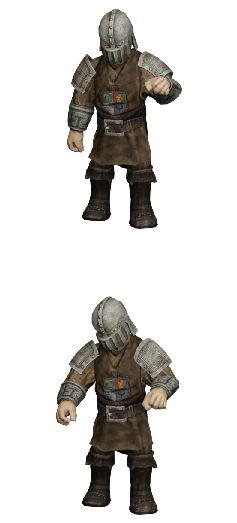}}~
    &
    \raisebox{-.0\height}
    {\includegraphics[width=0.19\textwidth,clip]{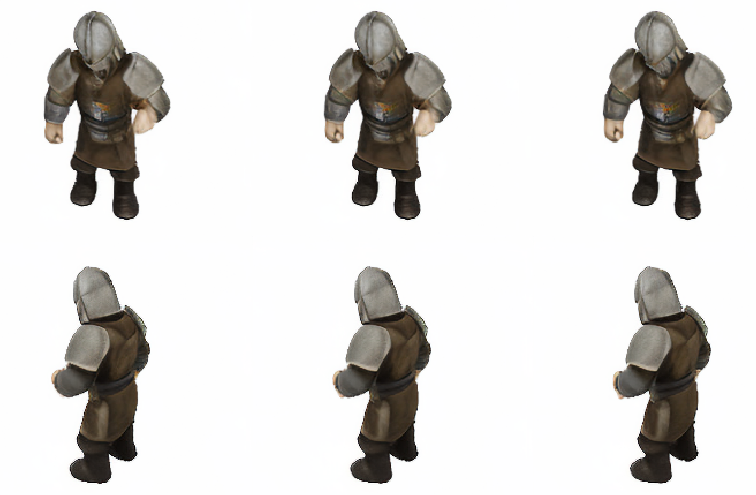}}~
    &
    \raisebox{.02\height}
    {\includegraphics[width=0.20\textwidth,clip]{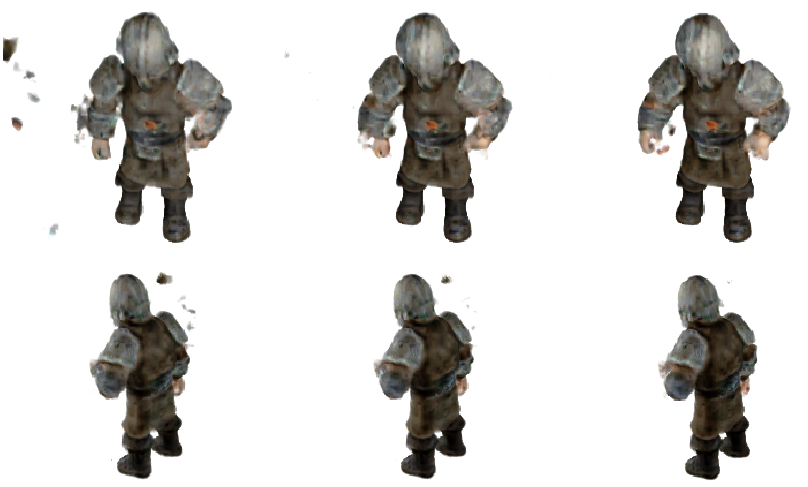}}

    \\
    Input & Ours & Consistent4D

    \end{tabular}
    \caption{\textcolor{black}{\textbf{Given only two input frames}, our method is able to generate smooth dynamics. For each case, we show three internal images from two novel views. }}
\label{fig:2input}
\end{figure}

\begin{figure}[t]

    \centering 
    \setlength{\tabcolsep}{0mm}{
    \begin{tabular}{c}

    {\includegraphics[width=0.49\textwidth,clip]{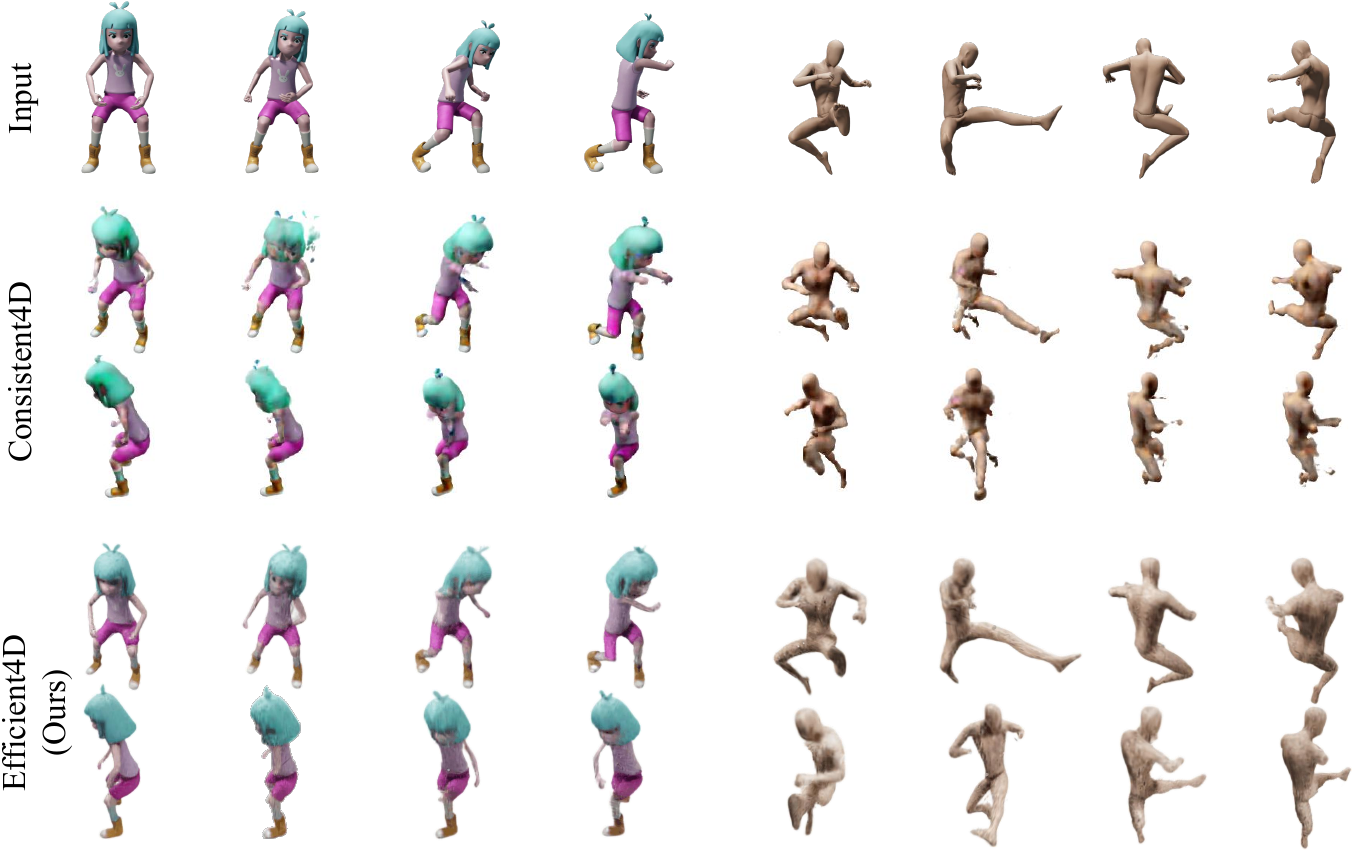}}
    
    \end{tabular}
    }
    
\caption{\textbf{Performance on cases including rotation.} We show two novel views for each case.}

\label{fig:rotation}
\end{figure}

We assessed our method's performance on two challenging cases: extremely sparse input and rotation dynamics.
In sparse case, the input comprising only two discrete frames. In such cases, we set $w = 0.25$ in equation~(\ref{eq:fuse_feature_2}). We also modified the code of Consistent4D~\cite{jiang2023consistent4d} to make it adapt to such case. As illustrated in Figure~\ref{fig:2input}, our approach successfully generates images featuring smooth motion and high spatiotemporal consistency. In contrast, Consistent4D fails to operate effectively under such conditions.
Consider a scenario where we seek 4D modeling for a static toy. While the toy can take different poses, it lacks autonomous movement, posing a challenge for continuous video capturing. In these instances, our method demonstrates effectiveness by requiring only a few key frames to produce dynamic content, thereby expanding the potential applications of the 4D generation task.

For another rotation case, we adopt the data introduced in \cite{yang2024diffusion}. As shown in the first row of Figure~\ref{fig:rotation}, the character will rotate its body in the monocular video input, which is complex and challenging.
However, our approach can also successfully infer the accurate geometry while Consistent4D tends to generate fragments. For example, from the Consistent4D results in Figure~\ref{fig:rotation}, we observe that the punching figure and the kicking man both suffer fragmented arms or legs.

\subsection{Ablation studies}
\label{sec:ablation}

\subsubsection{Ablation of image generation}

\begin{figure*}[t]
    \centering 
    \setlength{\tabcolsep}{0.0pt}
    \begin{tabular}{cc:c|c:c} 
    
    \rotatebox{90}{\makecell{Single\\ view}
    }
    &
    \raisebox{-.0\height}
    {\includegraphics[width=0.325\textwidth,clip]{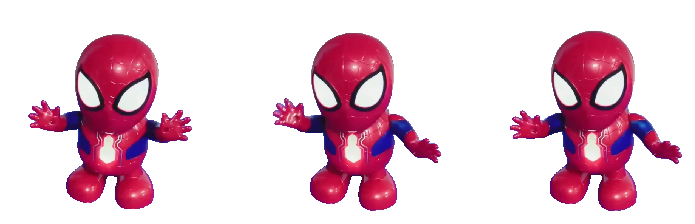}}~
    &
    \raisebox{-.0\height}
    {\includegraphics[width=0.13\textwidth,clip]{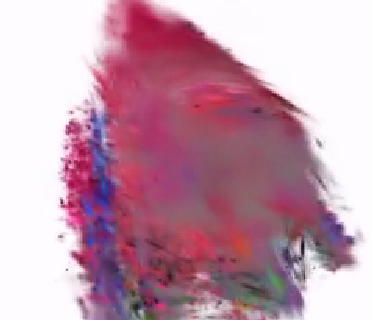}}
    &
    \raisebox{-.0\height}
    {\includegraphics[width=0.325\textwidth,clip]{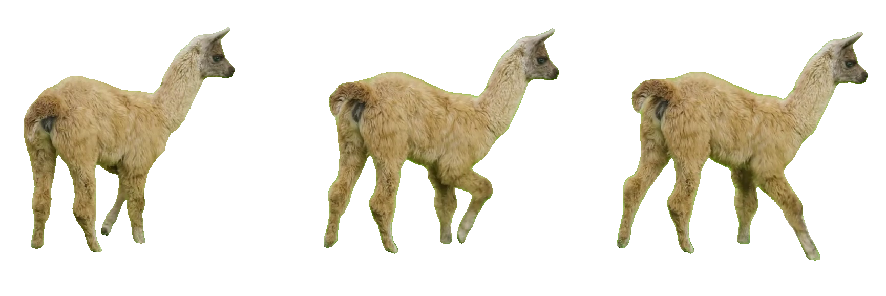}}~
    &
    \raisebox{-.0\height}
    {\includegraphics[width=0.13\textwidth,clip]{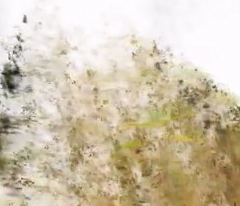}}
    \\
    \rotatebox{90}{\makecell{No\\ time-sync}}
    &
    \raisebox{-.0\height}
    {\includegraphics[width=0.325\textwidth,clip]{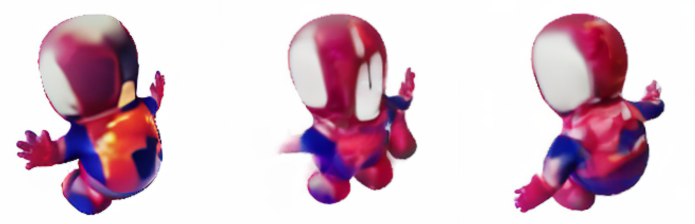}}~
    &
    \raisebox{-.0\height}
    {\includegraphics[width=0.13\textwidth,clip]{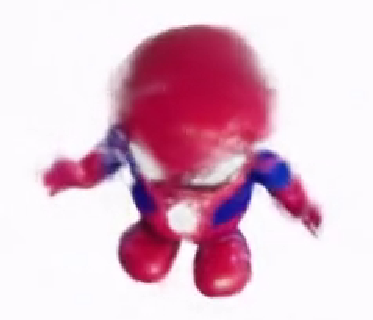}}
    &
    \raisebox{-.0\height}
    {\includegraphics[width=0.325\textwidth,clip]{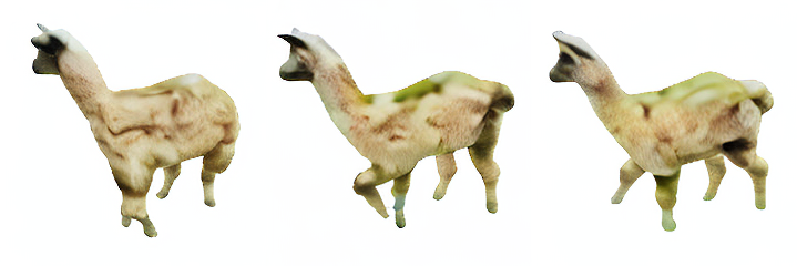}}~
    &
    \raisebox{-.0\height}
    {\includegraphics[width=0.13\textwidth,clip]{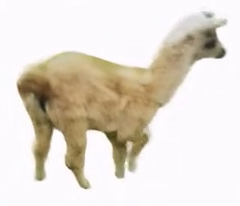}}
    \\
    \rotatebox{90}{\makecell{No \\ interp}}
    &
    \raisebox{-.1\height}
    {\includegraphics[width=0.325\textwidth,clip]{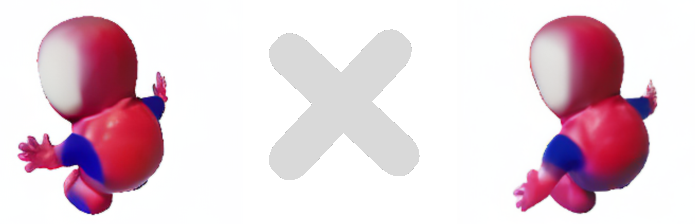}}~
    &
    \raisebox{-.1\height}
    {\includegraphics[width=0.13\textwidth,clip]{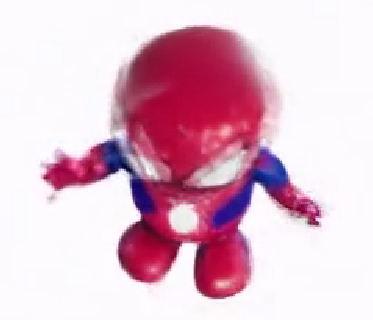}}
    &
    \raisebox{-.1\height}
    {\includegraphics[width=0.325\textwidth,clip]{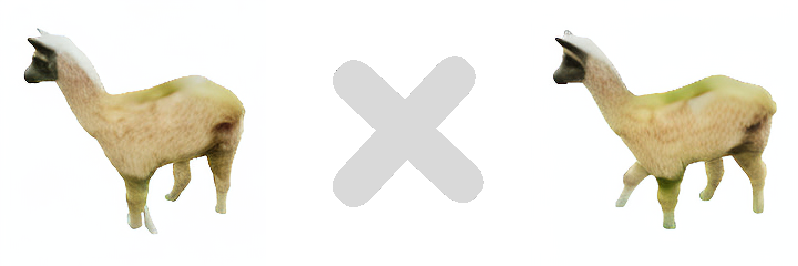}}~
    &
    \raisebox{-.1\height}
    {\includegraphics[width=0.13\textwidth,clip]{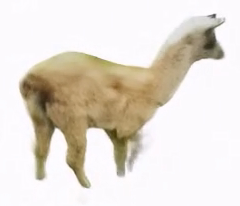}}
    \\
    \rotatebox{90}{\makecell{Full \\ setting}}
    &
    \raisebox{-.0\height}
    {\includegraphics[width=0.325\textwidth,clip]{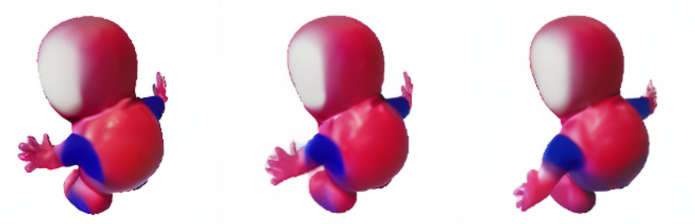}}~
    &
    \raisebox{-.0\height}
    {\includegraphics[width=0.13\textwidth,clip]{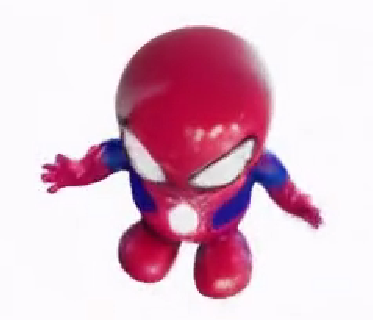}}
    &
    \raisebox{-.0\height}
    {\includegraphics[width=0.325\textwidth,clip]{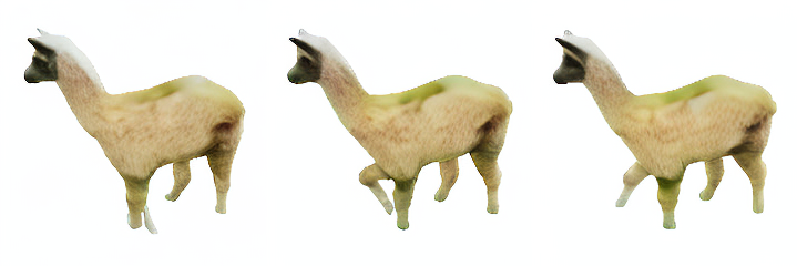}}~
    &
    \raisebox{-.0\height}
    {\includegraphics[width=0.13\textwidth,clip]{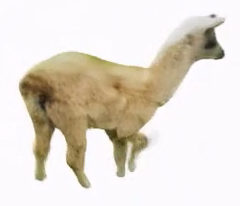}}
    \\
    & \begin{tabularx}{0.325\textwidth}{ >{\centering\arraybackslash}X >{\centering\arraybackslash}X >{\centering\arraybackslash}X}
      \small{$t_1$} &  \small{$t_2$} & \small{$t_3$}
    \end{tabularx}
    & \small{$t_2$} & \begin{tabularx}{0.325\textwidth}{ >{\centering\arraybackslash}X >{\centering\arraybackslash}X >{\centering\arraybackslash}X}
      \small{$t_1$} &  \small{$t_2$} & \small{$t_3$}
    \end{tabularx} & \small{$t_2$}
    \\
      & \small{Input/Generated images} & \small{Rendered} & \small{Input/Generated images} & \small{Rendered}
    \\

    \end{tabular}
    \caption{\textbf{Ablation study} on image generation, time-synchronous spatial volumes and frame interpolation. For each case, we show the input images in the first row and three novel-view images generated by SyncDreamer-T at different timestamps in the other rows, followed by one rendered image from the reconstructed 4DGS at the middle timestamp. 
    Each column refers to a specific timestep.
    For the row ``No interp'', there is no interpolated input image in the middle, thus leading to a vacancy
    at timestamp $t_2$.  } 
\label{fig:ablation_img}
\end{figure*}
\begin{table}[t]
\caption{\textcolor{black}{\textbf{Quantitative ablation study} on all our modules mesured on 4 sequences: alpaca, astronaut, rabbit and spiderman.}}
\label{tab:quan_ablation}

\begin{center}

\setlength{\tabcolsep}{0.5mm}{
\begin{tabular}{l|ccccc}

\hline

\hline

\hline
Method & CLIP $\uparrow$ & CLIP-T-f $\uparrow$ & CLIP-T-s $\uparrow$ & CLIP-T-b $\uparrow$ \\  
\hline

\hline

\hline

Single-view & 0.6684 & -  & - & - \\
No time-sync & 0.8270&0.9258&0.9197&0.8819\\
No interp & 0.8595&0.9409&0.9443&0.9336\\
Full setting & \textbf{0.8702}&\textbf{0.9689}&\textbf{0.9673}&\textbf{0.9684}\\

\hline

\hline

\hline

\end{tabular}
}
\end{center}

\end{table}

\begin{figure*}[t]

    \centering 
    \setlength{\tabcolsep}{0mm}{
    \begin{tabular}{c}

    {\includegraphics[width=0.95\textwidth,clip]{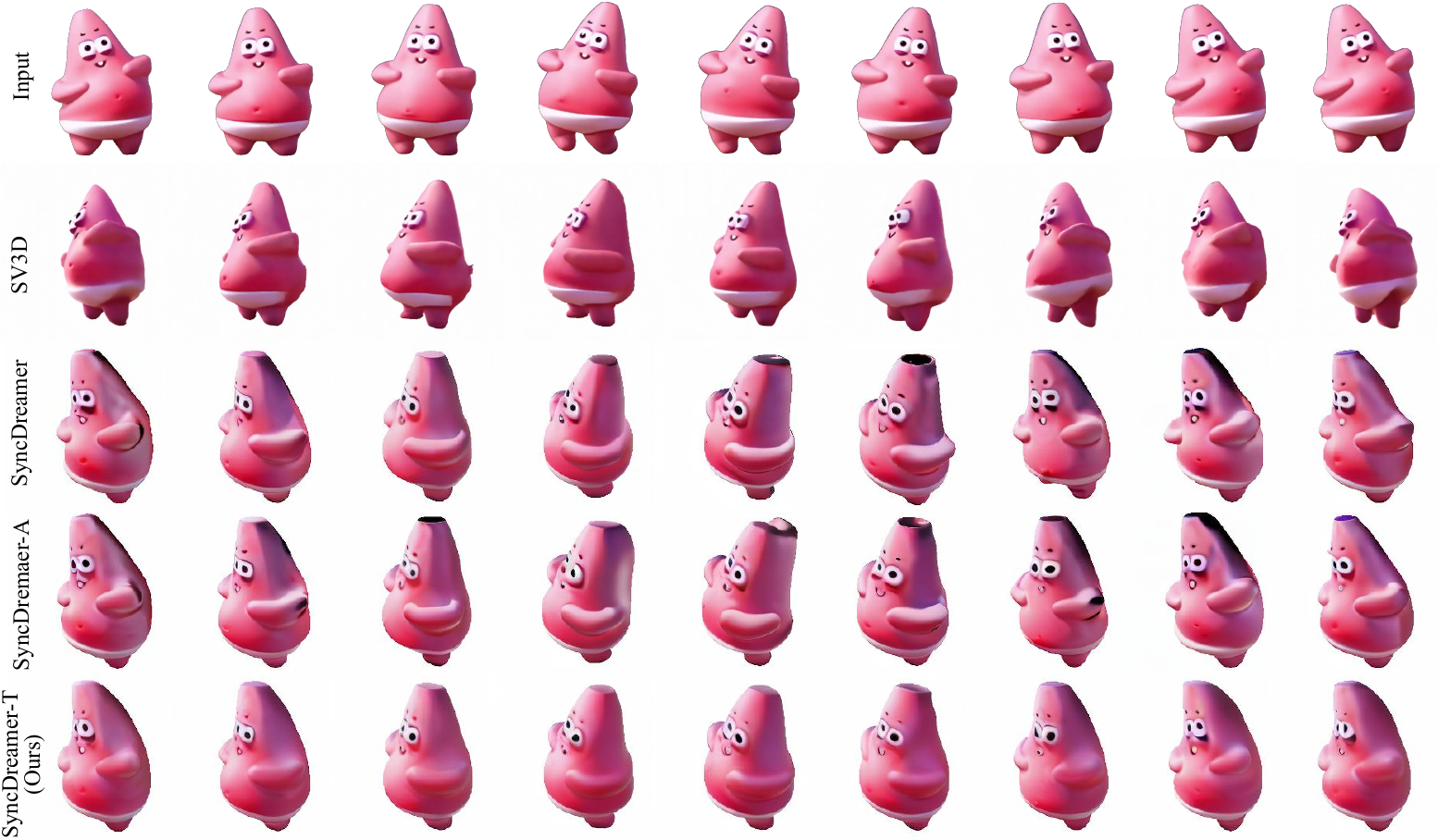}} \\
    {\includegraphics[width=0.95\textwidth,clip]{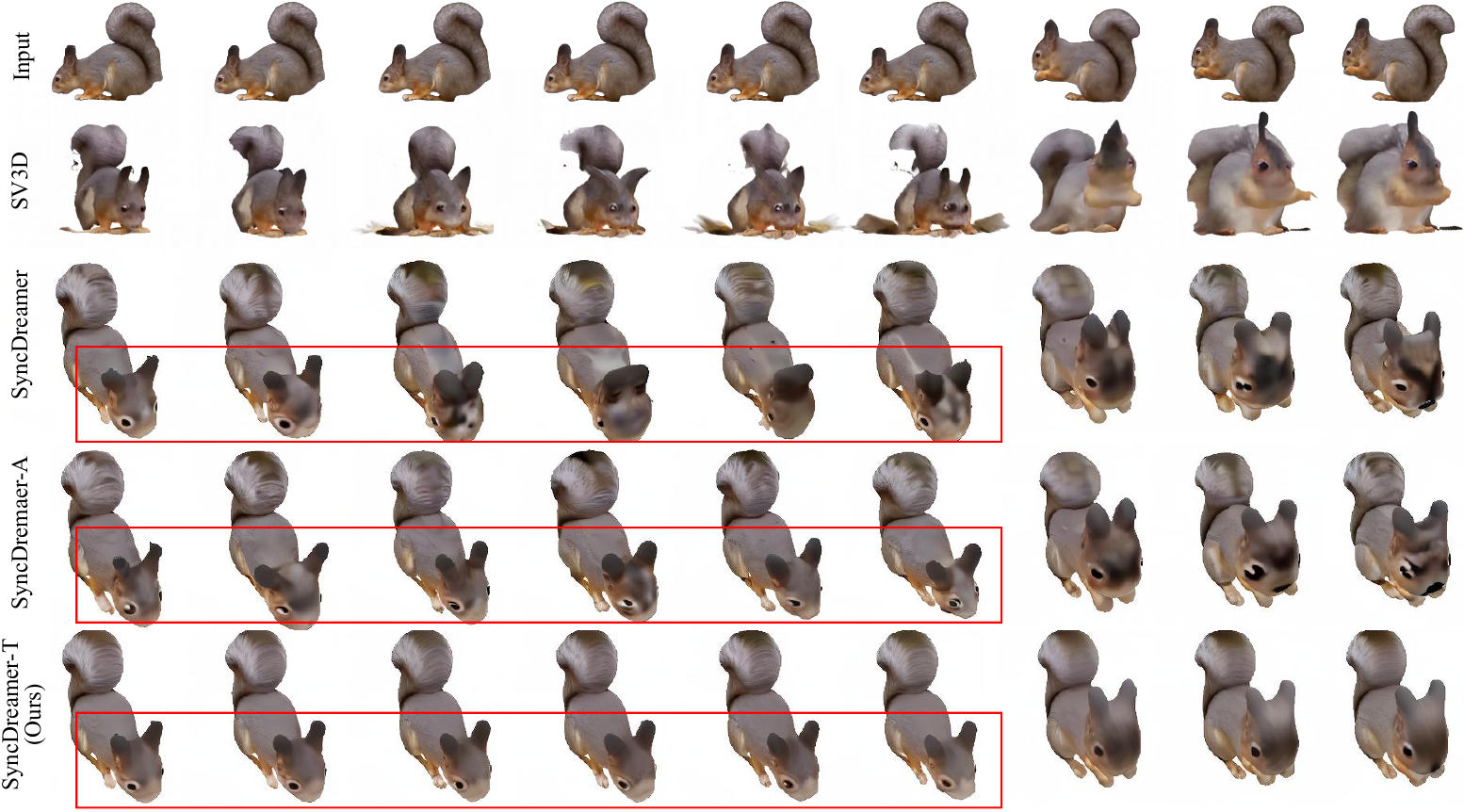}}
    
    \end{tabular}
    }
    
\caption{\textbf{Ablation study on the temporal synchronization.} We compare SV3D~\cite{voleti2024sv3d}, SyncDreamer~\cite{liu2023syncdreamer}, SyncDreamer with consistent self-attention~\cite{zhou2024storydiffusion} (SyncDreamer-A) and our proposed Syncdreamer-T. We show nine consecutive frames for each case to better evalute the temporal consistency.}

\label{fig:ablation_blending}
\end{figure*}

We first performed ablation studies to assess the influence of different components in our image generation stage. To better evaluate the image quality itself, we reconstruct the 4D Gaussian without SDS loss.
We compared our full method against three baseline settings: \textbf{(1)} Only input video are used for reconstruction; \textbf{(2)} Time-synchronous spatial volume is excluded; \textbf{(3)} Frame interpolation is excluded. Note that setting \textbf{(2)} is equivalent to the case where SyncDreamer-T is replaced with original SyncDreamer. The results corresponding to these settings can be found in Figure~\ref{fig:ablation_img}. 
Table~\ref{tab:quan_ablation} also gives quantitative evaluations. 
Beyond the above components, we also ablate the design of temporal synchronization in Figure~\ref{fig:ablation_blending}.

\noindent\textbf{Importance of synthetic data}
We compare our generated image matrix with utilizing only input video using the same 4D Gaussian representation model.
As shown in the first row of Figure~\ref{fig:ablation_img},  
when only relying on a single-view video, the model cannot produce any meaningful results for novel views. This indicates the importance of constructing proper training data.

\noindent\textbf{Effect of time-synchronous spatial volume}
We assess the impact of the time-synchronous spatial volume concept introduced in SyncDreamer, as depicted in the contrast of the second and last rows in Figure~\ref{fig:ablation_img}. Without time-synchronous spatial volume, the back aspects of the toy Spiderman exhibit inconsistencies, leading to distorted geometry. In contrast, the proposed time-synchronous spatial volume enhances both spatial and temporal consistency while preventing geometry collapse, resulting in more visually appealing image generation and higher CLIP-T scores in Table~\ref{tab:quan_ablation}.

\noindent\textbf{Effect of frame interpolation}
As illustrated in the contrast between the third and last rows in Figure~\ref{fig:ablation_img}, frame interpolation is effective in mitigating the blurring observed in the rendered image from novel views, thus also delivering higher CLIP-T scores in Table~\ref{tab:quan_ablation}. This is attributed to the low frame rates of the image matrix, which results in noticeable discontinuities.

\noindent\textbf{Ablation on temporal synchronization}
We also examine the design of temporal synchronization. 
In conceptual terms, directly applying image-to-3D methods to each frame independently is inferior due to intrinsic generative randomness. This can be validated in Figure~\ref{fig:ablation_blending} by comparing SV3D~\cite{voleti2024sv3d}, SyncDreamer~\cite{liu2023syncdreamer} and SyncDremaer-T(Ours).
The image-to-3D methods (SV3D and SyncDreamer) cannot generate a consistent shape or texture across frames.
Therefore, a proper temporal synchronization mechanism is necessary.
In Section~\ref{sec:image_gen}, we propose Sync-Dreamer-T, which blends the 3D volume feature. We also implement a consistent self-attention mechanism introduced in \cite{zhou2024storydiffusion} in SyncDreamer~\cite{liu2023syncdreamer}, resulting in Syncdreamer-A. Specifically, in each self-attention block that projects $Q_i, K_i, V_i$ corresponding to the image $I_i$, we randomly sample another image $I_j$'s $K_j, V_j$, and concatenate two $K$s and $V$s as the new $K_i$ and $V_i$.
The results in Figure~\ref{fig:ablation_blending} indicate that our method generates a more consistent texture and smoother motion than SyncDreamer-A.
We ascribe this to the following reason:
The cross-attended features may focus too much on the details
that its control on global texture and motion is weakened.

\subsubsection{Ablation of reconstruction stage}
\begin{figure}[t]

    \centering 
    \setlength{\tabcolsep}{0mm}{
    \begin{tabular}{c}

    {\includegraphics[width=0.49\textwidth,clip]{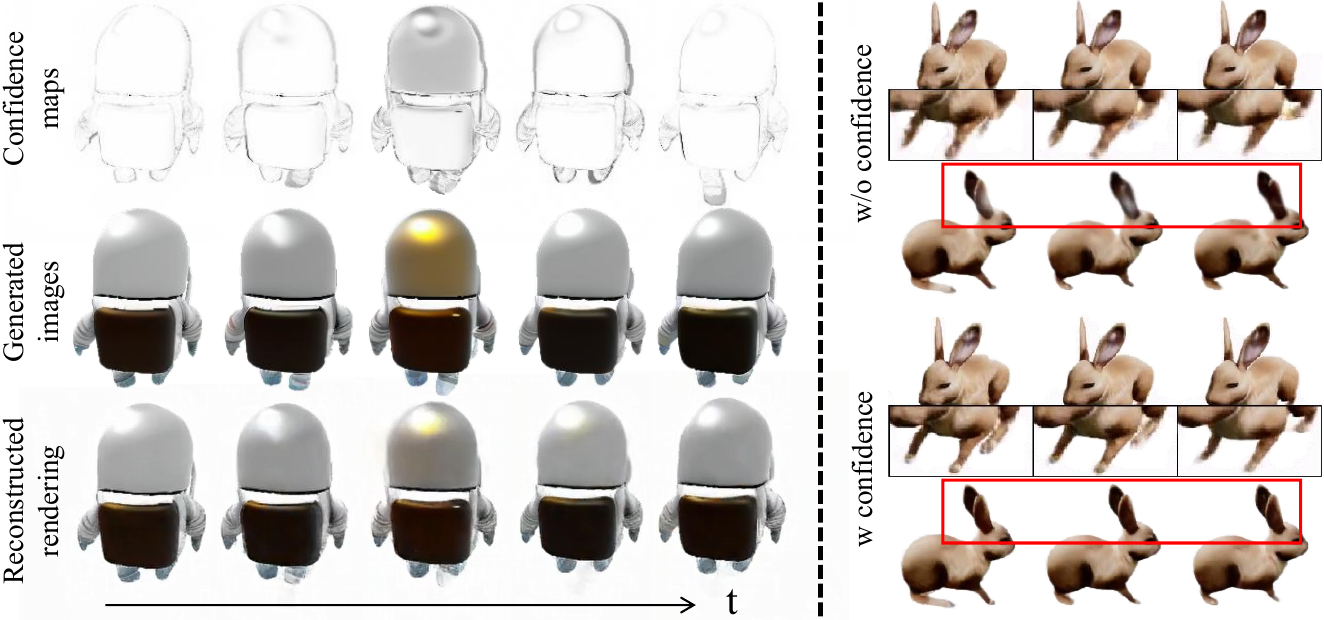}}
    
    \end{tabular}
    }
    
\caption{\textbf{Ablation study} on confidence maps. The images follow a chronological order from left to right.}

\label{fig:ablation_conf}
\end{figure}

\begin{figure}[t]

    \centering 
    \setlength{\tabcolsep}{0mm}{
    \begin{tabular}{c}

    {\includegraphics[width=0.49\textwidth,clip]{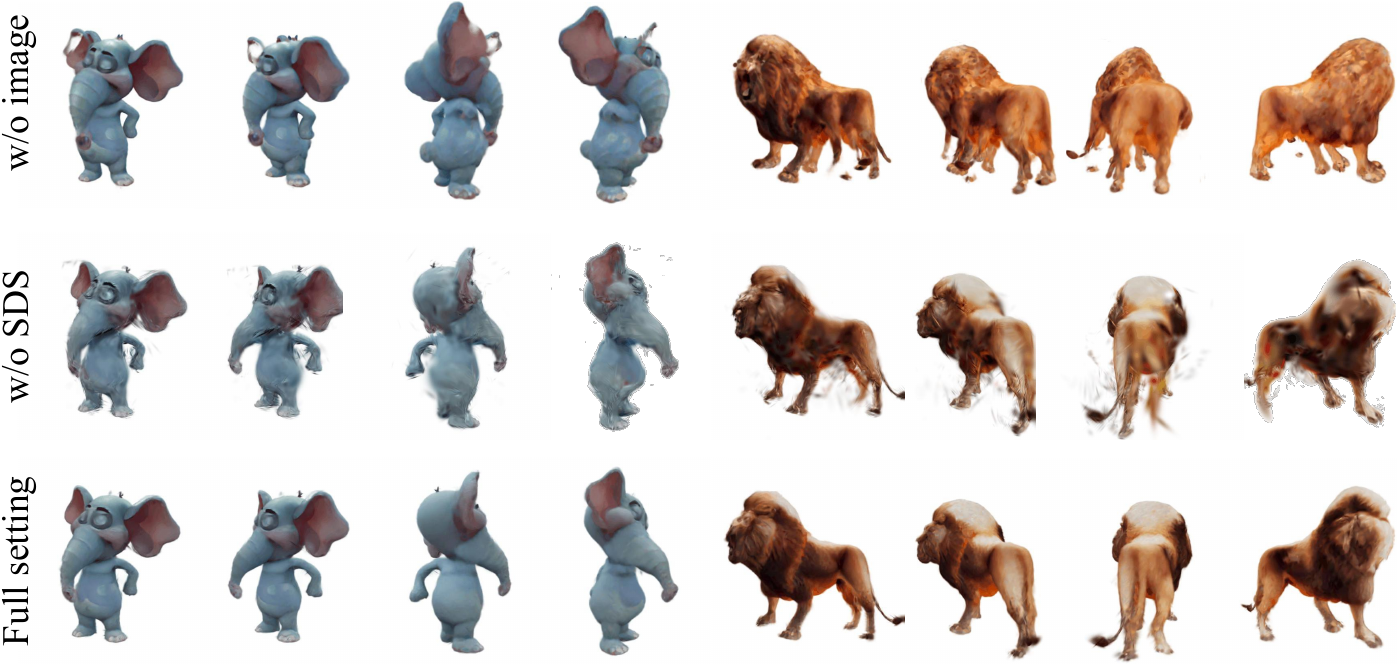}}
    
    \end{tabular}
    }
    
\caption{\textbf{Ablation study} on image supervision and SDS loss.}

\label{fig:ablation_sds}
\end{figure}

\begin{figure}[t]

    \centering 
    \setlength{\tabcolsep}{0mm}{
    \begin{tabular}{c}

    {\includegraphics[width=0.49\textwidth,clip]{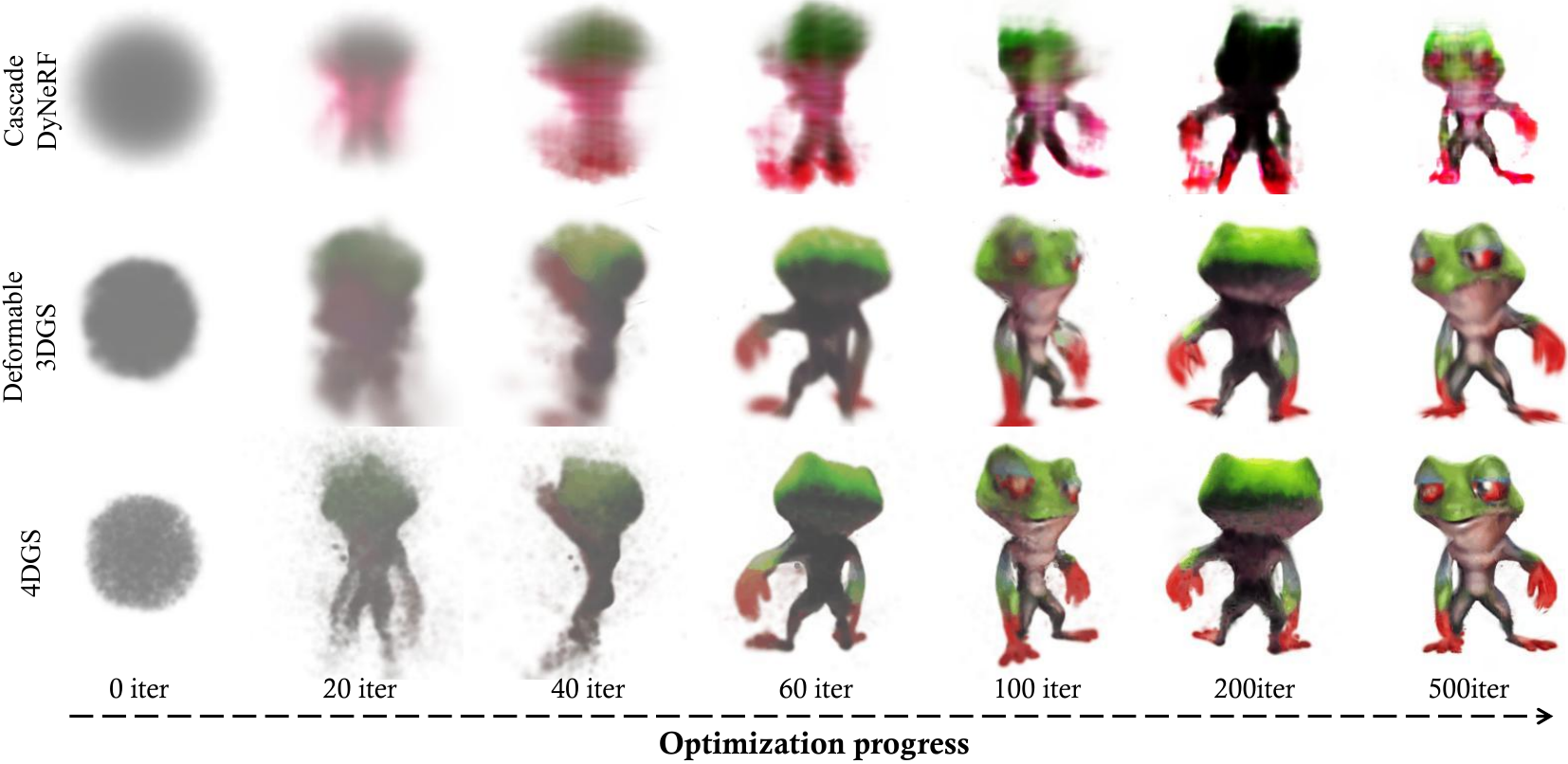}}
    
    \end{tabular}
    }
    
\caption{\textbf{Ablation study on the choice of 4D representation.} We apply the same optimization strategy to three 4D representations: Cascade-DyNeRF~\cite{jiang2023consistent4d}, Deformable-3DGS~\cite{wu20234dgaussians} and 4DGS~\cite{yang2023gs4d}. 4DGS features the fastest convergence speed.}

\label{fig:ablation_rep}
\end{figure}

In the reconstruction stage, we will study the effect of three components: \textbf{(1)} confidence map, \textbf{(2)} supervision of generated images, \textbf{(3)} SDS loss, and \textbf{(4)} choice of 4D representation. The visual results are shown in Figure~\ref{fig:ablation_conf}, \ref{fig:ablation_sds} and \ref{fig:ablation_rep}.

\noindent\textbf{Effect of confidence map}
The integration of a confidence weighted loss in our design serves as a strategy to mitigate training data noise. 
In the left of Figure~\ref{fig:ablation_conf}, we study the effect of confidence map when obvious inconsistency exists in the generated images. Our temporal smooth mechanism can support consistency for most areas, whilst a small fraction of lower-quality regions may introduce conflicting gradients in the reconstruction stage, thus hurting the overall quality.
To mitigate this, our confidence aware design comes into play for weakening the supervision from those inconsistent regions.
Since these weakly supervised regions only occupy a small proportion across all views, the missing information about texture or geometry  can be compensated by the redundancy of other views and generalization ability of 4D Gaussians.
As shown in the left of Figure~\ref{fig:ablation_conf}, the inclusion of confidence maps effectively reduces blurry and inconsistent rendering even when the generated images have plausible consistency, resulting in a significant overall improvement in quality and temporal smoothness.

\noindent\textbf{Importance of image supervision and SDS loss}
In Figure~\ref{fig:ablation_sds}, we compare the full setting of our method with baselines without image supervision (w/o image) or SDS loss (w/o SDS). The absence of anchored image supervision results in bad geometry, such as the hole in the elephant ears and multiple legs of the lion. This can be also attributed to the conflicts of SDS loss during prolonged optimization. In addition, due to the sparsity of the generated images, the novel views may not be reconstructed well, leading to blur and floaters. By contrast, our full method avoids the disadvantages of both, being able to render clean images with good geometry.

\noindent\textbf{Choice of 4D representation}
Our synthetic training data is versatile and supports the optimization of 4D representation models, such as Cascade-DyNeRF~\cite{jiang2023consistent4d} and Deformable-3DGS~\cite{wu20234dgaussians} used in other 4D generation works~\cite{jiang2023consistent4d, ren2023dreamgaussian4d}. In the comparison shown in Figure~\ref{fig:ablation_rep}, 4DGS~\cite{yang2023gs4d} converge in 500 iterations while the others do not. Also, thanks to the fast convergence speed of 4DGS, our method achieves significant efficiency advantage.

\section{Conclusion}
\label{sec:conclusion}

This study introduces a new framework, \model{}, designed for generating efficiently dynamic 4D objects seamlessly from monocular videos captured by a stationary camera. The \model{} consists of two main stages: first, generating consistent multi-view videos with spatial and temporal coherence, and second, rapidly producing 4D object reconstructions. Our approach, utilizing image supervision with lightly weighted SDS loss, significantly accelerates the generation process, achieving about 10 times faster speeds compared to previous works, while still delivering superior reconstruction and novel view synthesis results. Moreover, our model is effective in extremely sparse input scenarios, requiring only two available images, thereby expanding its application scope.


\section{Data Availability Statement}

The datasets used for evaluation and comparison are available at Consistent4D~\cite{jiang2023consistent4d} (\url{https://github.com/yanqinJiang/Consistent4D}), Animate124 \cite{zhao2023animate124} (\url{https://github.com/HeliosZhao/Animate124/tree/main/benchmark}) and Sketchfab~\cite{sketchfab} (\url{https://sketchfab.com/}).

\bibliographystyle{spbasic}      
\bibliography{main}   


\end{document}